\documentclass{article}

\usepackage{PRIMEarxiv}

\usepackage[utf8]{inputenc} % allow utf-8 input
\usepackage[T1]{fontenc}    % use 8-bit T1 fonts
\usepackage{hyperref}       % hyperlinks
\usepackage{url}            % simple URL typesetting
\usepackage{booktabs}       % professional-quality tables
\usepackage{amsfonts}       % blackboard math symbols
\usepackage{amsmath}        % math environments
\usepackage{nicefrac}       % compact symbols for 1/2, etc.
\usepackage{microtype}      % microtypography
\usepackage{fancyhdr}       % header
\usepackage{graphicx}       % graphics
\usepackage{xcolor}  
\graphicspath{{figures/}}     % organize your images and other figures under media/ folder

\title{scHFBiMamba: a bidirectional Mamba autoencoder for cardiac disease-state classification from single-nucleus RNA-seq data}

\author{
  Ugo Lomoio \\
  Department of Surgical and Medical Sciences \\
  University ``Magna Gr{\ae}cia'' of Catanzaro \\
  88100 Catanzaro, Italy \\
  \texttt{ugo.lomoio@unicz.it} \\
  \And
  Pietro Hiram Guzzi \thanks{Corresponding author.} \\
  Department of Surgical and Medical Sciences \\
  University ``Magna Gr{\ae}cia'' of Catanzaro \\
  88100 Catanzaro, Italy \\
  \texttt{hguzzi@unicz.it} \\
  \And
  Pierangelo Veltri \\
  DIMES \\
  University of Calabria \\
  88063 Rende, Italy \\
  \texttt{pierangelo.veltri@dimes.unical.it} \\
}

\begin{document}
\maketitle

\begin{abstract}
\textbf{Objective:} To evaluate whether a compact bidirectional state-space autoencoder can improve joint classification of species, cell type, and disease state from single-nucleus RNA-sequencing profiles in heart failure (HF) and related cardiac conditions, while providing biologically interpretable gene attributions.

\textbf{Methods:} Building on the dense denoising autoencoder (DAE) and multilayer perceptron (MLP) framework of Ruz Jurado et al., we developed \textbf{scHFBiMamba}, a framework comprising a bidirectional Mamba autoencoder (BiMambaAE) and an MLP classification head. BiMambaAE maps 16{,}545-gene single-nucleus expression profiles to 350-dimensional representations through learned latent tokens. The autoencoder contains 13.2M parameters, approximately \(14.6\times\) fewer than the dense DAE (192.9M). We compared six encoder--classifier configurations on a held-out test set of 38{,}275 cardiac nuclei, assessed nucleus-level differences using McNemar tests with Benjamini--Hochberg correction, and evaluated nine classifier variants on frozen BiMambaAE representations. Biological relevance was assessed by comparing SHAP-derived differentially explained genes with differentially expressed genes against curated HF-related Gene Ontology keywords.

\textbf{Results:} BiMambaAE with an MLP head achieved the highest aggregate F1 among the six configurations under both aggregate definitions: macro-F1 pooled over all 13 outputs was 0.966 versus 0.957 for DAE+MLP, and the arithmetic mean of the three category macro-F1 scores was 0.977 versus 0.970. BiMambaAE+MLP showed significantly higher cell-type and disease-state classification accuracy (Benjamini--Hochberg-adjusted \(p=7.1\times10^{-3}\) and \(p=4.8\times10^{-6}\), respectively), with a species F1 of 1.000. Among nine classifier variants, the MLP was a practical choice for prediction and gradient-based gene attribution. SHAP-derived differentially explained genes achieved a mean pathway-enrichment benchmark F1 of 0.105, compared with 0.053 for differentially expressed genes.

\textbf{Conclusion:} Compact bidirectional state-space autoencoders can outperform substantially larger dense autoencoders for joint cardiac transcriptomic classification while reducing parameter count. scHFBiMamba provides accurate multi-task prediction and hypothesis-generating gene-level interpretation, supporting further evaluation of bidirectional Mamba architectures for single-nucleus HF transcriptomics. Code is available at \url{https://github.com/UgoLomoio/scHFBiMamba}.
\end{abstract}


\section{Introduction}

Heart failure (HF) is a major cause of morbidity and mortality worldwide. Its clinical heterogeneity includes HF with reduced ejection fraction (HFrEF) and HF with preserved ejection fraction (HFpEF); related pressure-overload conditions, including aortic stenosis (AS), provide an additional context for studying cardiac remodeling \cite{savarese2023global}. Single-nucleus RNA sequencing (snRNA-seq) enables the characterization of cardiac disease at the level of individual cell types \cite{koenig2022single}. However, the dimensionality, sparsity, and cross-species heterogeneity of cardiac expression atlases complicate the identification of features that distinguish disease states.

Ruz Jurado et al.\ introduced a neural-network framework that classifies each cardiac nucleus by species (human or mouse), cell type (seven classes), and disease state (CTRL, AS, HFrEF, or HFpEF), using normalized expression profiles of 16{,}545 one-to-one orthologous genes \cite{jurado2025}. The framework combines a dense denoising autoencoder (DAE; 16545--5000--2400--350 encoder) with a multilayer perceptron (MLP; 795--230--105 hidden layers). Shapley additive explanations (SHAP) \cite{lundberg2017unified} are then used to identify \emph{differentially explained genes} (DXGs), whose model attributions differ between disease and control groups. The reported DXG analysis identified HF-related pathways that were not recovered using conventional differentially expressed genes (DEGs), motivating the integration of predictive modeling and attribution-based interpretation.

The parameter cost of the dense autoencoder motivates the evaluation of alternative architectures. The DAE contains $\approx$193M parameters, with much of this capacity allocated to fully connected input and reconstruction layers. Related biomedical applications have explored denoising convolutional autoencoders for ECG super-resolution \cite{lomoio2025design} and bidirectional Mamba-2 architectures for EEG super-resolution \cite{lomoio2026bidirectional}. Alongside attention-based Transformers \cite{vaswani2017attention}, including those used in single-cell models such as scGPT \cite{cui2024scgpt}, selective state-space models (SSMs) offer an alternative mechanism for sequence modeling. Mamba \cite{gu2023mamba} and Mamba-2 \cite{dao2024mamba2} support input-dependent sequence mixing with linear scaling in sequence length; bidirectional variants have also been investigated in biomedical signal reconstruction \cite{lomoio2026bidirectional}. These developments motivate testing whether a compact state-space autoencoder can improve the accuracy--parameter trade-off in cardiac snRNA-seq classification.

Single-cell foundation models provide a broader context for this question. Geneformer learns from rank-encoded gene sequences and was pretrained on $\sim$30M transcriptomes \cite{theodoris2023transfer}; scBERT applies BERT-style pretraining to single-cell expression data \cite{yang2022scbert}; the Universal Cell Embedding (UCE) model learns cross-species representations from 36M cells spanning eight species \cite{rosen2026uce}; and scFoundation uses a 100M-parameter architecture covering $\sim$20{,}000 genes and pretraining on over 50M profiles \cite{hao2024scfoundation}. These approaches target transferable representations across datasets and tasks. In contrast, the present study evaluates a task-specific architecture within an existing cardiac classification and DXG pipeline \cite{jurado2025}. The central question is whether this pipeline can retain predictive and attribution-based utility with substantially fewer autoencoder parameters. Pretraining cost, downstream adaptation cost, and attribution support are distinct considerations; neither model size nor pretraining alone precludes the use of a differentiable classifier or gene-level explanations.

Related approaches include probabilistic latent-variable models such as scVI \cite{lopez2018scvi} and several applications of state-space architectures to single-cell data. These include SC-MAMBA2 \cite{zhao2024scmamba2}, GeneMamba \cite{qi2025genemamba}, scMamba \cite{yuan2025scmamba}, and RegFormer \cite{hu2026regformer}. The architectural feature examined here is the placement of a bidirectional Mamba2 block after an initial expression bottleneck: sequence mixing acts on 32 \emph{learned latent tokens}, rather than directly on the 16{,}545 input genes. This reduces the sequence length processed by the SSM while retaining a differentiable mapping from gene expression to classifier outputs. The resulting 13.2M-parameter autoencoder is evaluated against the dense DAE using the same latent dimensionality and a shared downstream attribution workflow. The contribution is therefore a task-specific architectural and empirical comparison, rather than a claim of general superiority over pretrained single-cell models.

\textbf{scHFBiMamba} combines a bidirectional Mamba autoencoder (BiMambaAE) with a classifier for cardiac disease-state prediction from snRNA-seq data. The autoencoder projects expression profiles into 32 learned latent tokens, applies a bidirectional Mamba2 layer, and maps the resulting representation to the 350-dimensional latent space used by the DAE baseline. This common dimensionality supports comparison of downstream classifiers, although it does not imply identical latent geometry. The study makes three contributions:

\begin{itemize}
    \item \textbf{A compact state-space autoencoder.} BiMambaAE contains 13.2M parameters, approximately $14.6\times$ fewer than the dense DAE, and is evaluated for downstream species, cell-type, and disease-state classification.
    \item \textbf{Encoder--classifier and classifier-only comparisons.} Six encoder--classifier configurations (M0--M5) and nine classifier variants on frozen BiMambaAE representations are evaluated, including graph convolutional networks (GCN) \cite{kipf2017semi}, graph attention networks (GAT) \cite{velickovic2018graph}, and XGBoost \cite{chen2016xgboost}. The test set contains 38{,}275 nuclei from held-out biological samples, and pairwise differences in prediction correctness are assessed with McNemar's test \cite{mcnemar1947note}.
    \item \textbf{Attribution-based pathway evaluation.} SHAP-derived DXGs from the six main configurations are compared with expression-based DEGs using enrichment against curated HF-related Gene Ontology keywords.
\end{itemize}

\section{Methods}

\subsection{Dataset and preprocessing}

The analysis used the CVD dataset published by Ruz Jurado et al.\ \cite{jurado2025,cvd_dataset}, comprising 272{,}080 cardiac nuclei from human hearts and mouse models: 144{,}677 nuclei from healthy human hearts, 40{,}205 from healthy mouse hearts, 41{,}689 human and 3{,}594 mouse nuclei with AS, 19{,}330 human and 7{,}046 mouse nuclei with HFrEF, and 1{,}712 human and 13{,}827 mouse nuclei with HFpEF. The source preprocessing workflow used CellRanger (v7.0.0) \cite{zheng2017massively} with the hg38 and mm10 reference genomes. Barcodes with $<$300 or $>$6000 detected genes, or $>$5\% mitochondrial content, were excluded. Normalization, scaling, and integration followed a Seurat (v4.3.0) workflow \cite{hao2021integrated}. Mapping mouse genes to human one-to-one orthologs yielded 16{,}545 input features. CellTypist annotations \cite{dominguez2022cross} were manually curated into seven cell types: cardiomyocytes (CM), endothelial cells (EC), fibroblasts (FB), immune cells (IC), neuronal cells (NC), pericytes (PC), and smooth muscle cells (SMC).

Each nucleus has three categorical labels: species (2 classes), cell type (7 classes), and disease state (4 classes: CTRL, AS, HFrEF, HFpEF). These labels are concatenated into a 13-dimensional multi-hot target vector. Partitioning is performed at the biological-sample level, with joint stratification on the species$\times$cell-type$\times$disease label combination: training and validation follow a stratified 80/20 split over samples \cite{jurado2025}, and all nuclei from a held-out sample remain in the same partition. The test set comprises 38{,}275 nuclei from complete biological samples held out by species and disease state. Thus, test-set separation is defined at the biological-sample level, whereas the classification metrics reported here are calculated at the nucleus level. Test-set representations are computed with the frozen trained encoder; no part of the pipeline is fit on test-set expression.
% TODO (P3, author input needed): donor/animal counts and IDs per partition;
% whether multiple samples from one donor stay together in the same
% partition; reconciliation of the 272,080 total vs. the 38,275 test nuclei;
% exact training and validation set sizes.

\subsection{Baseline: dense DAE + MLP (M0)}

The reference architecture of Ruz Jurado et al.\ \cite{jurado2025} was reimplemented in PyTorch \cite{paszke2019pytorch}. The dense DAE encoder maps 16{,}545 normalized expression values through layers of 5000, 2400, and 350 units with ReLU activations. An asymmetric decoder reconstructs the input through hidden layers of 2200 and 5150 units. The MLP maps the 350-dimensional representation to 13 output scores through hidden layers of 795, 230, and 105 units with dropout ($p=0.1$); sigmoid activation yields output probabilities. This DAE+MLP configuration, denoted M0, is the replication baseline.

\subsection{Bidirectional Mamba autoencoder (BiMambaAE)}

\begin{figure}[t]
    \centering
    \includegraphics[width=\textwidth]{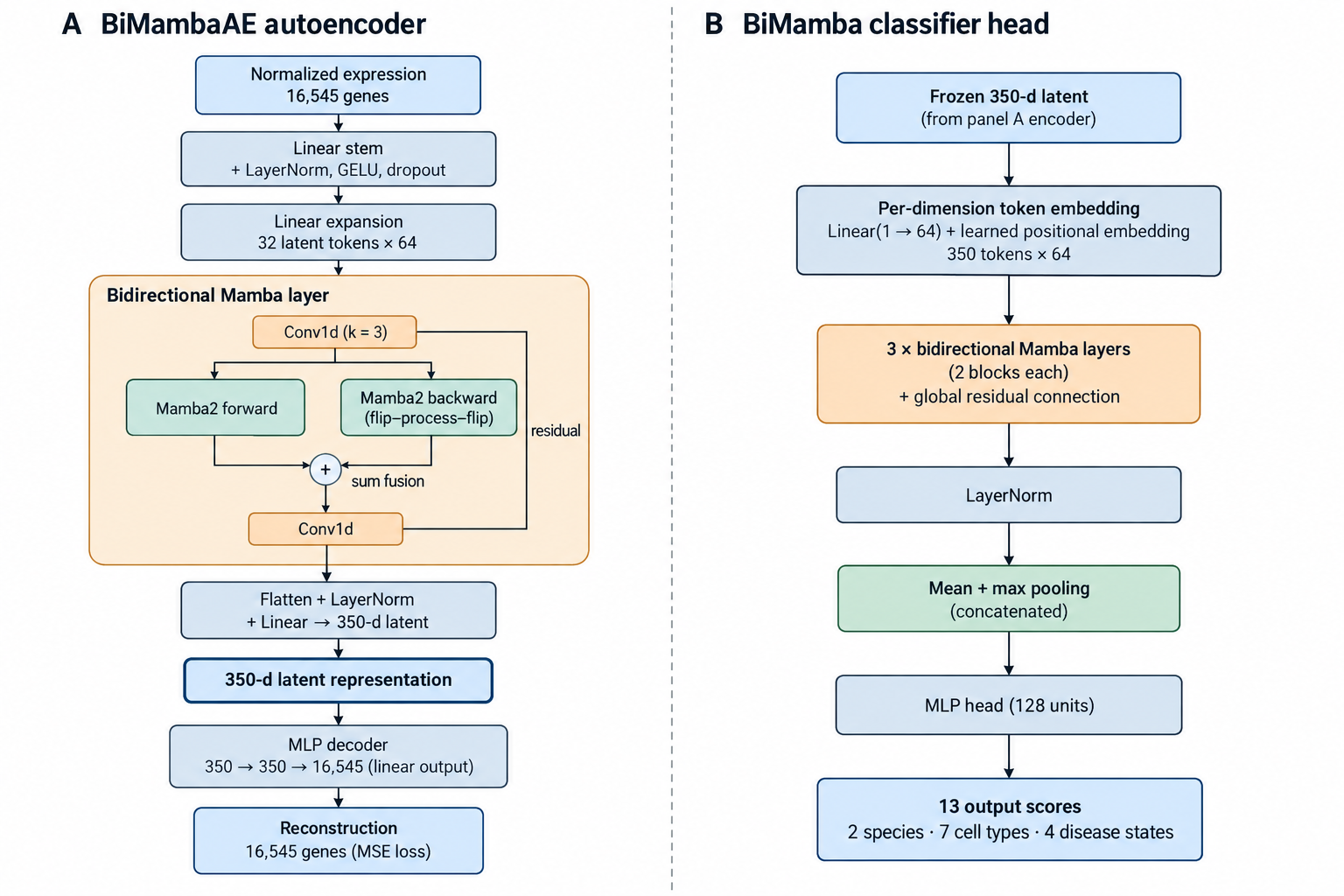}
    \caption{Architecture of the two main components. \textbf{Left:} the BiMambaAE autoencoder: a linear stem projects the 16{,}545-gene input to 350 dimensions, reshaped into 32 latent tokens of dimension 64; a bidirectional Mamba2 layer mixes the token sequence (the backward block processes the reversed sequence and its output is reversed back before summation); a projection maps the flattened tokens to the 350-dimensional latent vector; a two-layer MLP decoder reconstructs the input through a linear output layer. \textbf{Right:} the BiMamba classifier head: each latent dimension is embedded as a token with a learned positional embedding, processed by three bidirectional layers of two blocks each with a global residual connection, then mean- and max-pooled and mapped through a 128-unit head to 13 outputs (2 species, 7 cell types, 4 disease states).}
    \label{fig:architecture}
\end{figure}

\begin{figure}[t]
    \centering
    \includegraphics[width=\textwidth]{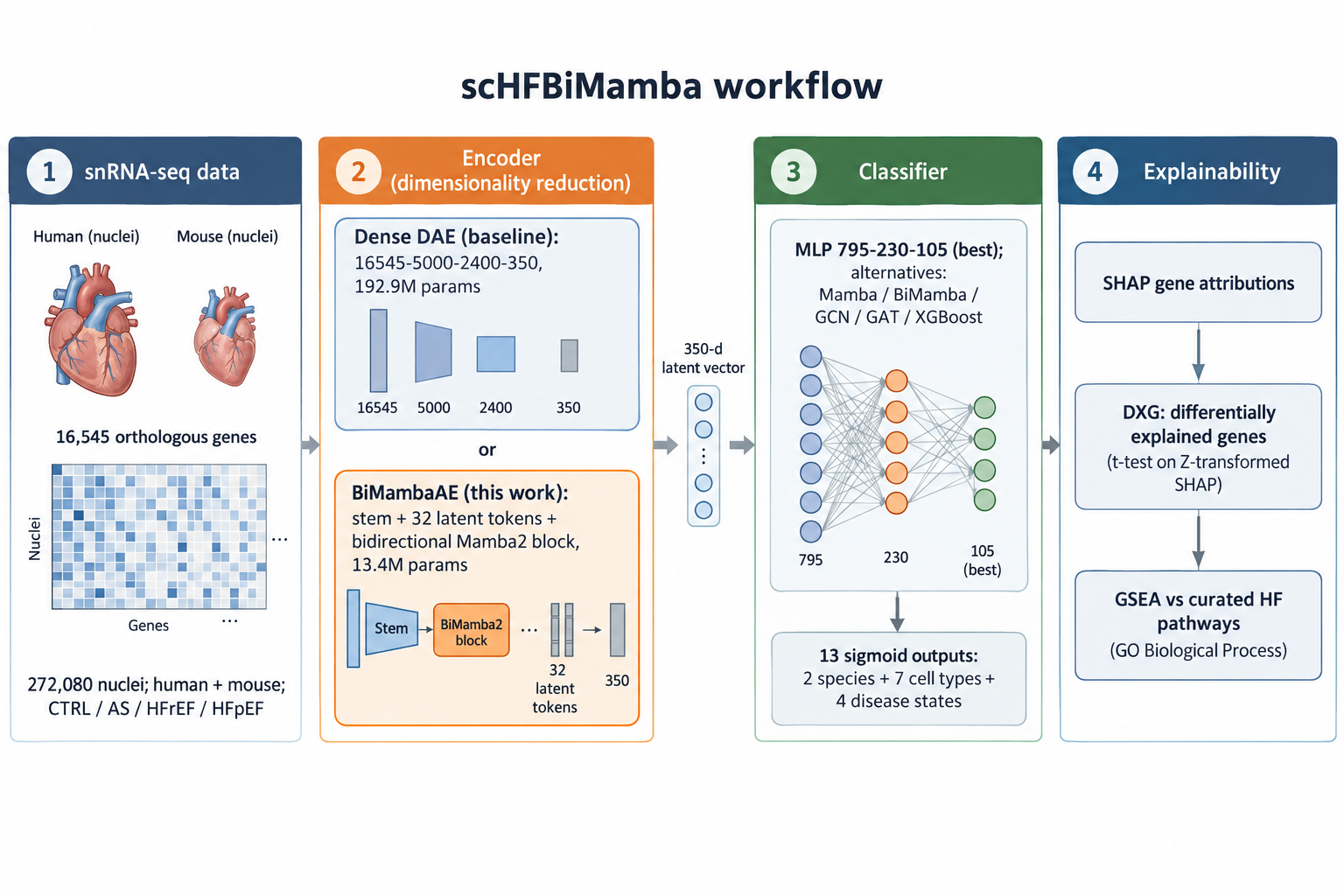}
    \caption{scHFBiMamba workflow. Normalized snRNA-seq profiles (16{,}545 orthologous genes; human and mouse nuclei across CTRL, AS, HFrEF, and HFpEF) are compressed by either the dense DAE baseline or the proposed BiMambaAE into a 350-dimensional latent space. A classifier head predicts 13 outputs spanning species, cell type, and disease state. SHAP attributions through the composite model feed differential explanation (DXG) analysis and GSEA-based pathway evaluation against curated HF-related GO terms.}
    \label{fig:workflow}
\end{figure}

BiMambaAE (Figs.~\ref{fig:workflow} and \ref{fig:architecture}) replaces the dense autoencoder with a compact architecture incorporating state-space sequence mixing. The encoder comprises: (i) a linear \emph{stem} projecting the input to 350 dimensions, followed by LayerNorm, GELU, and dropout ($p=0.1$); (ii) a linear expansion reshaped into 32 \emph{latent tokens} with dimension $d_{\text{model}}=64$; (iii) one bidirectional Mamba layer; and (iv) LayerNorm and a linear projection of the flattened tokens to a 350-dimensional latent vector. The decoder is a two-layer MLP (350$\rightarrow$350$\rightarrow$16545) with LayerNorm, GELU, and dropout.

The latent-token bottleneck limits the sequence length processed by the SSM to 32, rather than 16{,}545. Sequence mixing therefore operates on a compact learned representation, while the input and reconstruction projections account for much of the parameter budget. These tokens are learned features and are not assigned to individual genes or validated gene programs. Bidirectional scans propagate information in both directions along the token sequence; they do not make the representation permutation-invariant or confer a biological interpretation on token order.

The bidirectional Mamba layer applies an initial Conv1d operation with kernel size 3, followed by forward and backward Mamba2 blocks \cite{dao2024mamba2} ($d_{\text{state}}=16$, $d_{\text{conv}}=4$, expansion factor 2). The backward block processes the reversed sequence, and its output is reversed back to the original token order before summation (flip--process--flip), so that corresponding positions are aligned. A final convolution and an intra-layer residual connection link the initial convolution input to the final convolution output. This architecture enables input-dependent mixing among latent tokens in both directions. The complete autoencoder (encoder and decoder) contains 13{,}242{,}815 parameters (encoder 7{,}311{,}970; decoder 5{,}930{,}845), approximately $14.6\times$ fewer than the dense DAE (192{,}903{,}395); the deployed inference path (BiMambaAE encoder plus MLP head, decoder discarded) contains 7{,}799{,}728 parameters, and the complete M3 model contains 13{,}730{,}573 (Supplementary File S3). Both autoencoders use mean-squared-error reconstruction loss; no input corruption (masking or noise) was applied during autoencoder training, and the ``denoising'' designation is retained from the source study. The BiMambaAE decoder ends in a linear (identity-activation) output layer. However, their decoders and optimization settings also differ, so the comparison does not isolate the effect of the Mamba block alone.

\subsection{Classifier heads}

Three classifier heads were evaluated on each encoder's 350-dimensional representation: (i) the reference MLP (795--230--105--13); (ii) a bidirectional Mamba classifier that embeds each latent dimension as a token using a linear projection and a learned positional embedding, applies three bidirectional Mamba layers of two blocks each with a global residual connection, and concatenates mean and max pooling before a 128-unit classification head; and (iii) a unidirectional Mamba variant. Each of the 13 outputs is a raw logit trained with \texttt{BCEWithLogitsLoss}; at inference, predictions are formed by argmax within each category group, so exactly one label per category (species, cell type, disease state) is assigned per nucleus. The separate classifier ablation used frozen autoencoder representations and additionally included XGBoost (three multiclass models, one per target category; 300 trees, maximum depth 6, learning rate 0.1) \cite{chen2016xgboost}; a two-layer GCN and a two-layer GAT (hidden dimension 64, two attention heads) \cite{kipf2017semi,velickovic2018graph} operating transductively on a Euclidean $k$-nearest-neighbor graph ($k=15$) built on the pooled train/validation/test latents, with test labels masked: the graph loss is computed on training nodes, model selection uses validation nodes, and test nodes receive no supervision; and BiMamba variants using class-weighted binary cross-entropy, attention pooling, or both.

\subsection{Training protocol}

Training was sequential, not joint: each autoencoder was first trained with mean-squared-error (MSE) reconstruction loss, then frozen; its latents were cached for the full dataset, and the classifier was trained separately on the cached latents. Neural classifiers used binary cross-entropy (BCE) with logits, with per-class positive weights where indicated. The dense DAE and neural classifiers used Adam \cite{kingma2015adam} (learning rate $10^{-3}$, batch size 1024, maximum 500 epochs); BiMambaAE used AdamW \cite{loshchilov2019decoupled} (learning rate $10^{-4}$, weight decay $10^{-6}$, batch size 32, maximum 100 epochs). Neural-network runs used ReduceLROnPlateau scheduling (factor 0.1, patience 25, minimum learning rate $10^{-7}$), early stopping (patience 50), automatic mixed precision, and a fixed seed (1234). The six main configurations were M0, DAE+MLP; M1, DAE+Mamba (unidirectional); M2, DAE+BiMamba; M3, BiMambaAE+MLP; M4, BiMambaAE+Mamba; and M5, BiMambaAE+BiMamba. No training wall-clock records were retained in the results bundle, so no training-speed claims are made.

\subsection{Attribution, DXG, DEG, and pathway evaluation}

Gene-level attributions were calculated using SHAP GradientExplainer \cite{lundberg2017unified} on the composite encoder--classifier model. GradientExplainer uses expected gradients to approximate SHAP values; its outputs are not exact Shapley values. The composite model's outputs are sigmoid probabilities, so attributions refer to the probability scale. The attribution procedure used a background of 1000 training cells, balanced sampling of 200 test cells per cell type, \texttt{nsamples=100}, and seed 1111; mitochondrial-gene attributions were zeroed post hoc. Differentiability of both the encoder and classifier permits attribution of classifier outputs to the original gene-expression inputs. This property motivated the use of the MLP as the default classifier in the DXG workflow.

Following Ruz Jurado et al.\ \cite{jurado2025}, attributions were Z-transformed within each cell and compared between disease and control groups within each species$\times$cell-type stratum using Student's $t$-test. Multiple testing was controlled using Benjamini--Hochberg (BH) false discovery rate (FDR) adjustment \cite{benjamini1995controlling}. DXGs were filtered using adjusted $p<0.05$, SHAP magnitude $\geq0.01$, cube-root fold-change $\geq0.1$, and mean expression $\geq0.1$. DEGs were identified using the Wilcoxon rank-sum test on expression values, with adjusted $p<0.05$, $|\log_2\mathrm{FC}|\geq0.25$, and detection in $\geq10\%$ of cells in either group.

Preranked gene set enrichment analysis (GSEA) \cite{subramanian2005gene} was performed with GSEApy \cite{fang2023gseapy} against GO Biological Process 2023 \cite{aleksander2023gene}, using 1000 permutations and gene-set sizes of 10--500, with significance at FDR $q\leq0.05$. Genes were ranked by $\mathrm{sign(FC)}\cdot\log_{10}(|\mathrm{FC}|/p_{\text{bin}}+1)$, where $p_{\text{bin}}=0.1$ if the BH-adjusted $p<0.05$ and $p_{\text{bin}}=1$ otherwise. Significant terms were matched case-insensitively (substring matching) to the curated, cell-type-specific reference keywords in Supplementary Table 8 of the source study. For each contrast and method, true positives (TP) are matched significant terms, false positives (FP) are unmatched significant terms, and false negatives (FN) are reference keywords not recovered among the significant terms; the per-contrast F1 is $2\,\mathrm{TP}/(2\,\mathrm{TP}+\mathrm{FP}+\mathrm{FN})$. This evaluation measures recovery of a curated keyword-based benchmark, rather than exhaustive biological ground truth. All 42 possible species$\times$cell-type$\times$condition contrasts, their availability flags, and per-model F1 values are listed in Supplementary File S1.

\subsection{Statistical comparison and efficiency benchmarking}

Two aggregate F1 statistics are reported side by side throughout. The \emph{pooled-13 macro-F1} is the macro average over all 13 outputs, weighting each output class equally; the \emph{mean-category F1} is the arithmetic mean of the three category macro-F1 scores (cell type, species, disease state). The two differ because the categories contain 7, 2, and 4 classes; both are recomputed at full precision in Supplementary File S2.

Test-set predictions were compared pairwise using continuity-corrected McNemar tests \cite{mcnemar1947note} for species, cell type, and disease state, with BH adjustment across 45 comparisons \cite{benjamini1995controlling}. These tests assess paired differences in nucleus-level prediction correctness, not differences in macro-F1, and treat nuclei as independent; within-sample dependence is acknowledged in the Limitations. Per-class and macro-averaged receiver operating characteristic (ROC) curves and areas under the ROC and precision--recall curves were calculated from prediction scores. Inference time was summarized as mean$\pm$SD over 50 batches of 1024 cells after 10 warm-up batches, with device synchronization around each timed GPU iteration. Parameter counts and estimated floating-point operations (FLOPs) were also recorded. FLOP conventions are stated in the Table~\ref{tab:efficiency} caption: dense layers are estimated as $2\times$parameters, and SSMs as $2\times\text{params}\times\text{sequence length}$ (32 latent tokens for BiMambaAE, 350 for the SSM classifier heads); graph-network estimates depend on the number of edges. XGBoost was benchmarked on CPU, with tree-node counts reported in place of neural-network parameters. These measurements characterize the reported implementations and execution settings rather than a hardware-independent ranking.
% TODO (P6, author input needed): benchmark hardware (GPU/CPU model), CUDA,
% PyTorch, and mamba-ssm versions for the reported timings.

\section{Results}

\subsection{Replication of the DAE+MLP baseline}

The PyTorch implementation of the reference architecture (M0) achieved performance broadly consistent with the source study \cite{jurado2025}. On the held-out test set of 38{,}275 nuclei, the pooled-13 aggregate F1 was 0.957 (mean-category F1 0.970), with category-level F1 values of 0.933 for cell type, 0.9996 for species, and 0.978 for disease state (Table~\ref{tab:ablation}, Fig.~\ref{fig:rates}). The source study reported a validation F1 of $\approx$0.95. Because these results concern different evaluation partitions, their agreement supports the use of M0 as a replication comparator but does not establish exact reproducibility.

\subsection{BiMambaAE improves classification with a smaller autoencoder}

\begin{figure}[t]
    \centering
    \includegraphics[width=\textwidth]{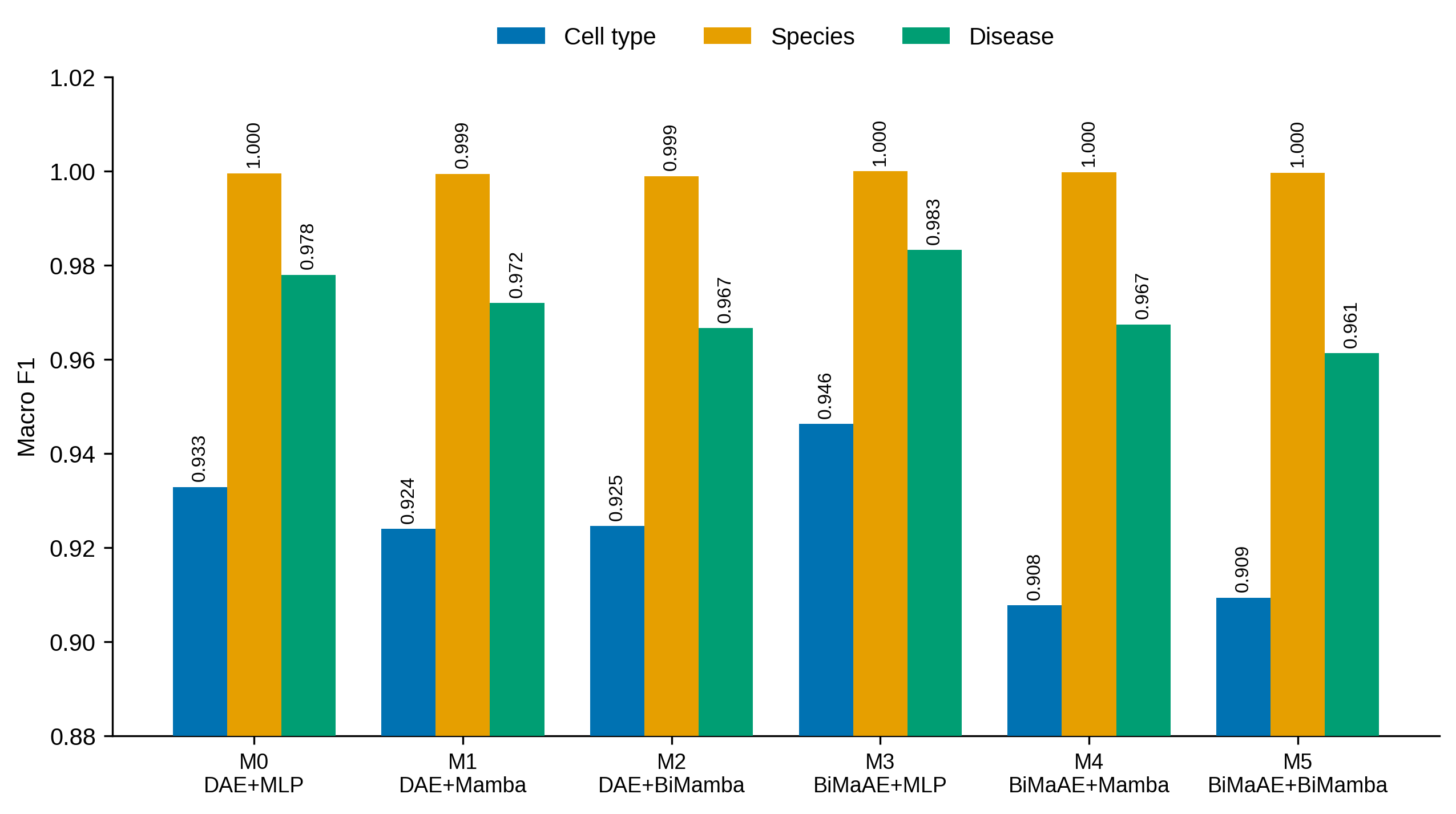}
    \caption{Macro-F1 per category (cell type, species, disease) for the six ablation models on the held-out test set. M3 (BiMambaAE+MLP) achieves the best cell-type (0.946) and disease (0.983) F1 and perfect species classification.}
    \label{fig:rates}
\end{figure}

\begin{figure}[t]
    \centering
    \begin{minipage}{0.49\textwidth}
        \includegraphics[width=\textwidth]{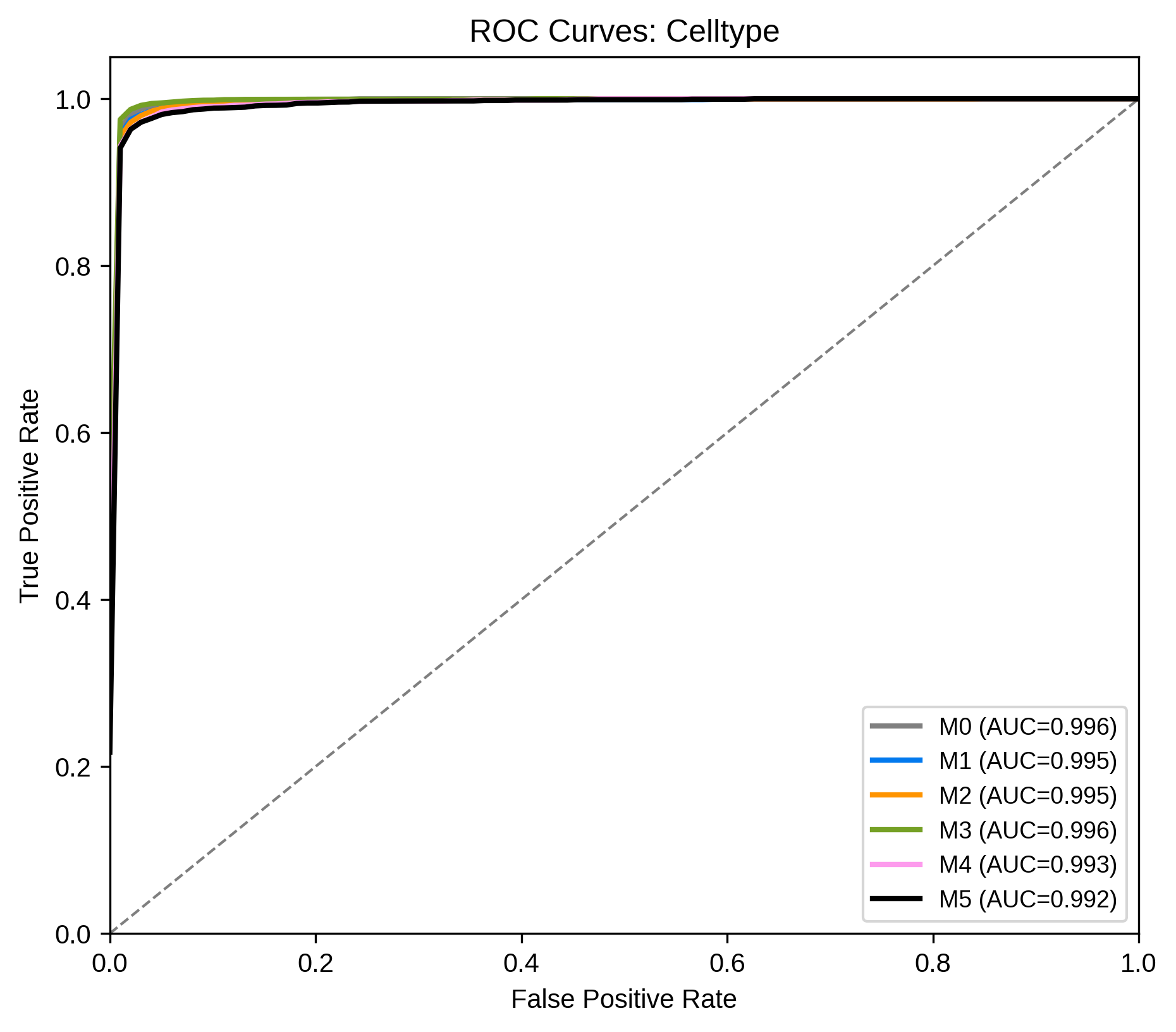}
    \end{minipage}\hfill
    \begin{minipage}{0.49\textwidth}
        \includegraphics[width=\textwidth]{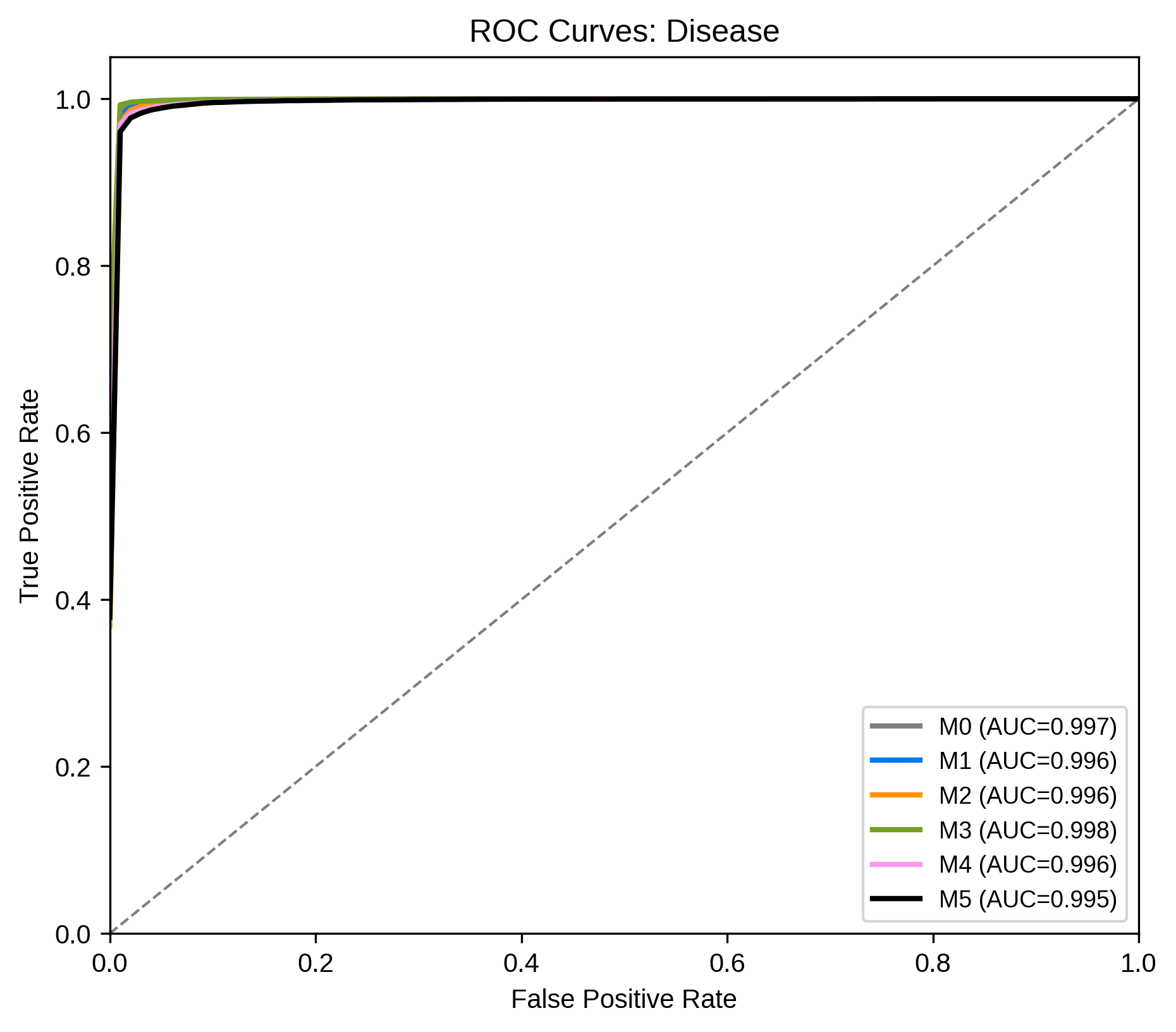}
    \end{minipage}
    \caption{Macro-averaged ROC curves on the held-out test set for cell-type (left) and disease-state (right) classification. M3 attains the highest macro ROC-AUC in both categories (0.9965 and 0.9981, respectively).}
    \label{fig:roc}
\end{figure}

Replacing the dense DAE with BiMambaAE while retaining the MLP head (M3) produced the highest aggregate F1 among the six main configurations under both definitions: 0.966 pooled-13 and 0.977 mean-category, with category-level F1 values of 0.946 for cell type, 1.000 for species, and 0.983 for disease state. The corresponding macro ROC-AUC values were 0.9965, 1.0000, and 0.9981 (Table~\ref{tab:ablation}, Figs.~\ref{fig:rates} and \ref{fig:roc}). M3 was the only configuration with significantly higher nucleus-level classification accuracy than M0 in all three categories: BH-adjusted McNemar $p=7.1\times10^{-3}$ for cell type, $p=4.8\times10^{-6}$ for disease state, and $p=7.4\times10^{-4}$ for species. The remaining configurations had lower cell-type and disease-state accuracy than M0 (all adjusted $p\leq1.2\times10^{-3}$; Table~\ref{tab:ablation}). With the dense DAE, replacing the MLP with Mamba heads (M1 and M2) did not improve performance. With BiMambaAE, the Mamba heads (M4 and M5) also underperformed the MLP (M3 vs.\ M4/M5: adjusted $p\leq6.1\times10^{-33}$ for cell type and $p\leq1.0\times10^{-50}$ for disease state). These comparisons favor the BiMambaAE+MLP configuration under the reported training settings; they do not isolate architecture from optimization effects.

\begin{figure}[t]
    \centering
    \includegraphics[width=0.75\textwidth]{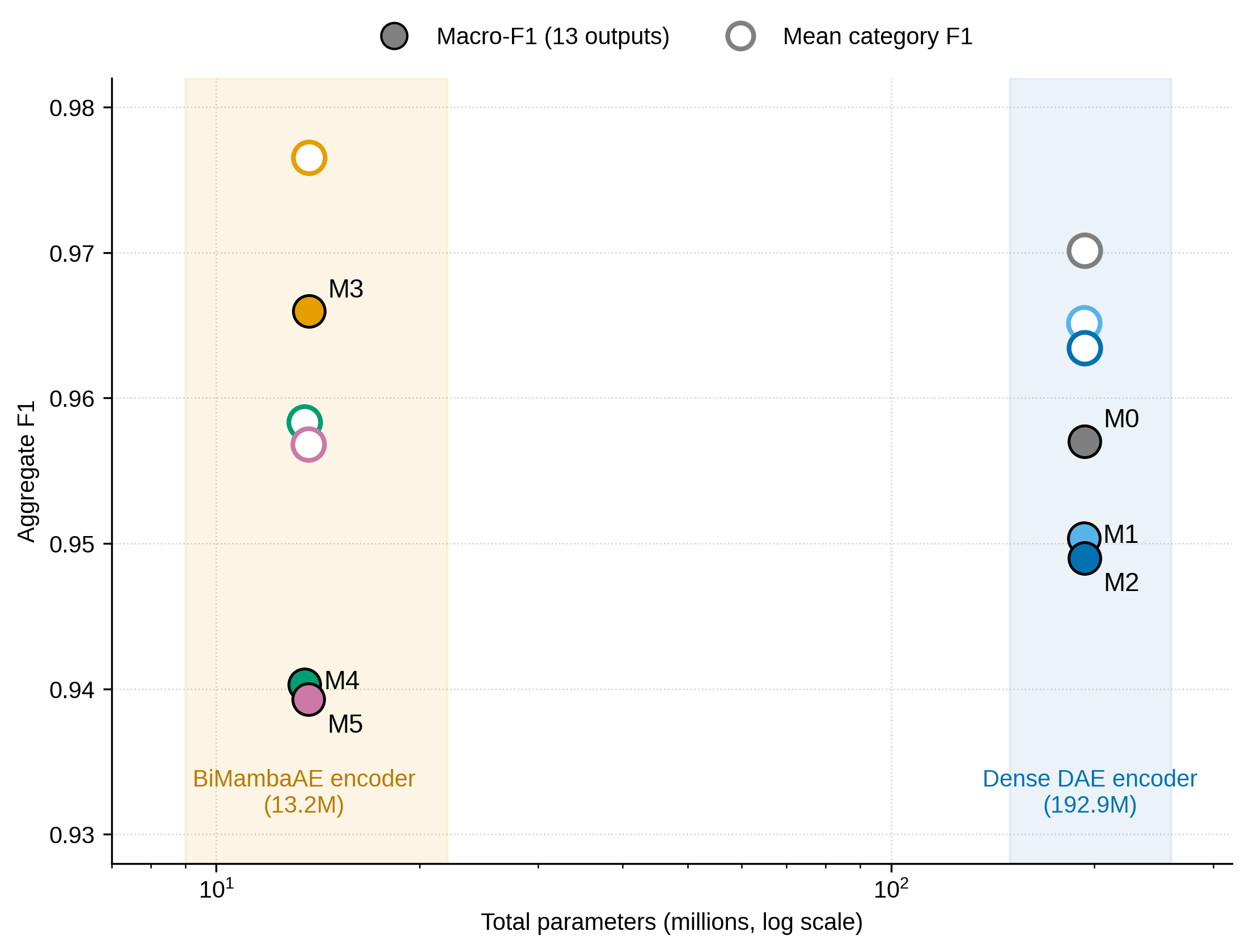}
    \caption{Classification performance versus model size. Total parameter count (log scale) is plotted against both aggregate F1 statistics for the six configurations (filled markers: pooled-13 macro-F1; open markers: mean-category F1). BiMambaAE configurations use a 13.2M-parameter autoencoder (orange shading), compared with 192.9M for the dense DAE (blue shading). M3 has the highest aggregate F1 under both definitions among these configurations.}
    \label{fig:params}
\end{figure}

The parameter comparison favors BiMambaAE (Fig.~\ref{fig:params}). The complete autoencoder contains 13.2M parameters, compared with 192.9M for the dense DAE, corresponding to a $14.6\times$ reduction. Including the classifier, M3 contains $\approx$13.7M parameters and achieves higher classification performance than the $\approx$193M-parameter DAE-based configurations.

\begin{table}[t]
\centering
\caption{Encoder--classifier comparison on the held-out test set of 38{,}275 nuclei. Category-level macro-F1 and both aggregate F1 statistics are shown: ``F1 mean (13)'' is the macro-F1 pooled over all 13 outputs (each class weighted equally); ``F1 mean (cat.)'' is the arithmetic mean of the three category macro-F1 scores. Both aggregates are computed at full precision from the stored per-class scores (Supplementary File S2); displayed category values are rounded, so their arithmetic mean can differ in the last digit. The highest value in each column is in \textbf{bold}; M0 comparator values are underlined where shown. BH-adjusted McNemar tests compare nucleus-level prediction correctness, not F1: M3 exceeds M0 in all three categories (cell type $p=7.1\times10^{-3}$, disease $p=4.8\times10^{-6}$, species $p=7.4\times10^{-4}$). M1, M2, M4, and M5 have lower cell-type and disease-state accuracy than M0 (all adjusted $p\leq1.2\times10^{-3}$); species differences are not significant for M4/M5.}
\label{tab:ablation}
\begin{tabular}{lllccccc}
\toprule
\textbf{Model} & \textbf{Encoder} & \textbf{Classifier} & \textbf{F1 cell type} & \textbf{F1 species} & \textbf{F1 disease} & \textbf{F1 mean (13)} & \textbf{F1 mean (cat.)} \\
\midrule
M0 & Dense DAE & MLP \cite{jurado2025} & \underline{0.9329} & 0.9996 & \underline{0.9780} & \underline{0.9570} & \underline{0.9702} \\
M1 & Dense DAE & Mamba (uni) & 0.9240 & 0.9994 & 0.9720 & 0.9504 & 0.9652 \\
M2 & Dense DAE & BiMamba & 0.9246 & 0.9990 & 0.9667 & 0.9490 & 0.9634 \\
\midrule
\textbf{M3} & \textbf{BiMambaAE} & \textbf{MLP} & \textbf{0.9463} & \textbf{1.0000} & \textbf{0.9833} & \textbf{0.9660} & \textbf{0.9765} \\
M4 & BiMambaAE & Mamba (uni) & 0.9078 & 0.9998 & 0.9674 & 0.9403 & 0.9584 \\
M5 & BiMambaAE & BiMamba & 0.9094 & 0.9997 & 0.9613 & 0.9393 & 0.9568 \\
\bottomrule
\end{tabular}
\end{table}

\subsection{Classifier ablation on frozen autoencoder latent spaces}

Nine classifier variants were trained on frozen DAE and BiMambaAE representations (Table~\ref{tab:classifier_dae}, Table~\ref{tab:classifier_mamba}). The provenance of every row is as follows. In Table~\ref{tab:classifier_dae} (frozen DAE representations), the Mamba and BiMamba (plain BCE) rows are the M1 and M2 main-experiment configurations, reused for $k=1$, seed 1234 (same encoder, same train split, same seed); the remaining rows come from the dedicated classifier-ablation run on DAE latents. In Table~\ref{tab:classifier_mamba} (frozen BiMambaAE representations), the MLP, Mamba, and BiMamba (plain BCE) rows are the M3, M4, and M5 main-experiment configurations, reused under the same conditions; the remaining rows come from the dedicated run on BiMambaAE latents. Because M3 shares the same frozen encoder and sequential training protocol across both experiments, its scores are common to the two tables.

An earlier version of the classifier comparison evaluated the BiMambaAE-trained SSM checkpoints on DAE latents (a checkpoint--latent mismatch). That run collapsed to degenerate majority-class predictions (mean F1 $\approx$0.11, with bit-identical species F1 across arms, the signature of a constant majority-class predictor); it is disclosed here because it motivated the per-latent-space presentation. Class imbalance alone is therefore not established as the explanation for the collapsed run.

With DAE embeddings, XGBoost achieved the highest mean-category F1 (0.972), followed closely by the MLP (0.970); both exceeded the transductive graph classifiers (GCN 0.913, GAT 0.897).

Table~\ref{tab:classifier_mamba} compares classifiers trained on frozen 350-dimensional BiMamba-AE representations. A simple MLP achieved the best performance across all tasks (mean-category F1 $=0.9765$), followed by XGBoost ($0.9721$) and the $k$NN-graph models GCN ($0.9671$) and GAT ($0.9650$). Species classification was nearly saturated for all models (F1 $\geq 0.9994$), so differences were driven mainly by cell type (F1 $0.8981$--$0.9463$) and disease. Sequence-based heads (Mamba, BiMamba) performed slightly worse (mean-category F1 $0.9489$--$0.9604$), and attention did not help. The weighted loss improved BiMamba on cell type ($0.9094\rightarrow0.9220$), but the gains remained below the MLP. These results indicate that the frozen embeddings are highly linearly and non-linearly separable, so lightweight classifiers suffice. These results identify sensitivity to the training and aggregation choices in this experiment; they do not establish label imbalance as the sole cause of the collapsed run or imply that all SSM classifiers require the same remedies.
% TODO (P2, author input needed): archive the aggregated_metrics.json of the
% new tab:classifier_mamba run with the revision; the XGBoost/GCN/GAT/
% weighted/attention rows on BiMambaAE latents were verified only for
% internal consistency (mean-of-3 matches the displayed category values).
% Optionally re-run the classifier ablation with code_patches.diff patch 3.

The MLP also offered low classifier-only inference latency in the reported benchmark (Table~\ref{tab:efficiency}): 0.154~ms per 1024-cell batch, compared with 192~ms for BiMamba and 51.3~ms for CPU XGBoost, corresponding to approximate ratios of 1250$\times$ and 330$\times$, respectively. BiMambaAE was about 2.2× faster than the dense DAE (3.23±0.95 vs. 7.01±0.04 ms) and used about $14.6\times$ fewer parameters (13.2M vs. 192.9M). Its estimated computational cost was also lower: 31.45M FLOPs per nucleus, because the SSM processes only the 32 latent tokens rather than the 16{,}545 genes, compared with 385.81M for the dense DAE ($2\times$parameters). This is consistent with the $14.6\times$ parameter reduction and the observed 2.2× speedup. The dense DAE showed lower variability in latency (±0.04 vs. ±0.95 ms).

These timings exclude the autoencoder, and the XGBoost comparison involves a different execution device. The MLP additionally supports gradient propagation through the encoder to the original gene inputs. Tree ensembles can be explained using other attribution methods, but they do not provide this differentiable composite path. The MLP was therefore selected as the default head for the DXG workflow.

\begin{table}[t]
\centering
\caption{Classifier comparison using frozen 350-dimensional DAE representations. Category-level macro-F1 and the mean-category F1 (arithmetic mean of the three category macro-F1 scores) are shown for the held-out test set (single run, seed 1234). The pooled-13 macro-F1 for the main configurations (M0--M2) is reported in Table~\ref{tab:ablation}. The highest value in each column is in \textbf{bold}.}
\label{tab:classifier_dae}
\begin{tabular}{lcccc}
\toprule
\textbf{Classifier} & \textbf{F1 cell type} & \textbf{F1 species} & \textbf{F1 disease} & \textbf{Mean cat.\ F1} \\
\midrule
XGBoost & \textbf{0.9381} & 0.9994 & \textbf{0.9787} & \textbf{0.9721} \\
MLP & 0.9362 & \textbf{0.9996} & 0.9748 & 0.9702 \\
GCN ($k$NN graph) & 0.8094 & 0.9993 & 0.9309 & 0.9132 \\
GAT ($k$NN graph) & 0.7721 & 0.9994 & 0.9194 & 0.8970 \\
Mamba (plain BCE) & 0.9240 & 0.9994 & 0.9720 & 0.9504 \\
BiMamba (plain BCE) & 0.9246 & 0.9990 & 0.9667 & 0.9490 \\
BiMamba + attention & 0.9229 & 0.9991 & 0.9649 & 0.9623 \\
BiMamba + weighted loss & 0.8915 & 0.9987 & 0.9634 & 0.9512 \\
BiMamba + weighted + attention & 0.8841 & 0.9983 & 0.9578 & 0.9468 \\
\bottomrule
\end{tabular}
\end{table}

\begin{table}[t]
\centering
\caption{Classifier comparison using frozen 350-dimensional BiMamba-AE representations. Category-level macro-F1 and the mean-category F1 (arithmetic mean of the three category macro-F1 scores) are shown for the held-out test set (single run, seed 1234). The pooled-13 macro-F1 for the main configurations (M3--M5) is reported in Table~\ref{tab:ablation}. The highest value in each column is in \textbf{bold}.}
\label{tab:classifier_mamba}
\begin{tabular}{lcccc}
\toprule
\textbf{Classifier} & \textbf{F1 cell type} & \textbf{F1 species} & \textbf{F1 disease} & \textbf{Mean cat.\ F1} \\
\midrule
MLP & \textbf{0.9463} & \textbf{1.0000} & \textbf{0.9833} & \textbf{0.9765} \\
XGBoost & 0.9410 & 0.9998 & 0.9755 & 0.9721 \\
GCN ($k$NN graph) & 0.9276 & 1.0000 & 0.9738 & 0.9671 \\
GAT ($k$NN graph) & 0.9247 & 1.0000 & 0.9704 & 0.9650 \\
Mamba (plain BCE) & 0.9078 & 0.9998 & 0.9674 & 0.9584 \\
BiMamba (plain BCE) & 0.9094 & 0.9997 & 0.9613 & 0.9568 \\
BiMamba + attention & 0.8981 & 0.9994 & 0.9489 & 0.9489 \\
BiMamba + weighted loss & 0.9220 & 0.9996 & 0.9595 & 0.9604 \\
BiMamba + weighted + attention & 0.8989 & 0.9994 & 0.9500 & 0.9494 \\
\bottomrule
\end{tabular}
\end{table}

\begin{table}[t]
\centering
\caption{Inference efficiency and model complexity. Inference time is mean $\pm$ SD per 1024-cell batch (50 batches after 10 warm-up, device-synchronized GPU iterations). XGBoost was benchmarked on CPU (parameter count reports total tree nodes; FLOPs not meaningful). FLOP conventions: dense layers estimated as $2\times$parameters; SSMs as $2\times$parameters$\times$sequence length (32 latent tokens for BiMambaAE, 350 for the SSM classifier heads); GCN/GAT FLOPs scale with graph edges (full-graph transductive pass normalized per 1024 cells). GCN/GAT parameter counts refer to the trained hidden-64 configuration (23{,}309 and 23{,}386 parameters); their FLOPs and timings were measured with the hidden-128 benchmark configuration (46{,}605 and 46{,}746 parameters), a mismatch disclosed here and in the Limitations.}
\label{tab:efficiency}
\begin{tabular}{lccc}
\toprule
\textbf{Component} & \textbf{Inference time (ms)} & \textbf{Parameters} & \textbf{FLOPs} \\
\midrule
\multicolumn{4}{l}{\emph{Autoencoders (encoder and decoder)}} \\
Dense DAE & 7.01$\pm$0.04 & 192.9M & 385.81M \\
BiMambaAE & 3.23$\pm$0.95 & \textbf{13.2M} & 31.45M \\
\midrule
\multicolumn{4}{l}{\emph{Classifiers (350-d latent input)}} \\
MLP & 0.15 $\pm$ 0.02 & 487.6K & 0.98M \\
Mamba & 98.04 $\pm$ 0.23 & 258.5K & 181.0M \\
BiMamba & 192.01 $\pm$ 0.21 & 424.7K & 297.3M \\
GCN & 0.07 $\pm$ 0.00 & 23.3K & 26.2B \\
GAT & 0.78 $\pm$ 0.00 & 23.4K & 25.4B \\
XGBoost (CPU) & 51.30 $\pm$ 1.90 & 286.6K & -- \\
\bottomrule
\end{tabular}
\end{table}

\subsection{Latent-space structure}

\begin{figure}[t]
    \centering
    \begin{minipage}{0.49\textwidth}
        \includegraphics[width=\textwidth]{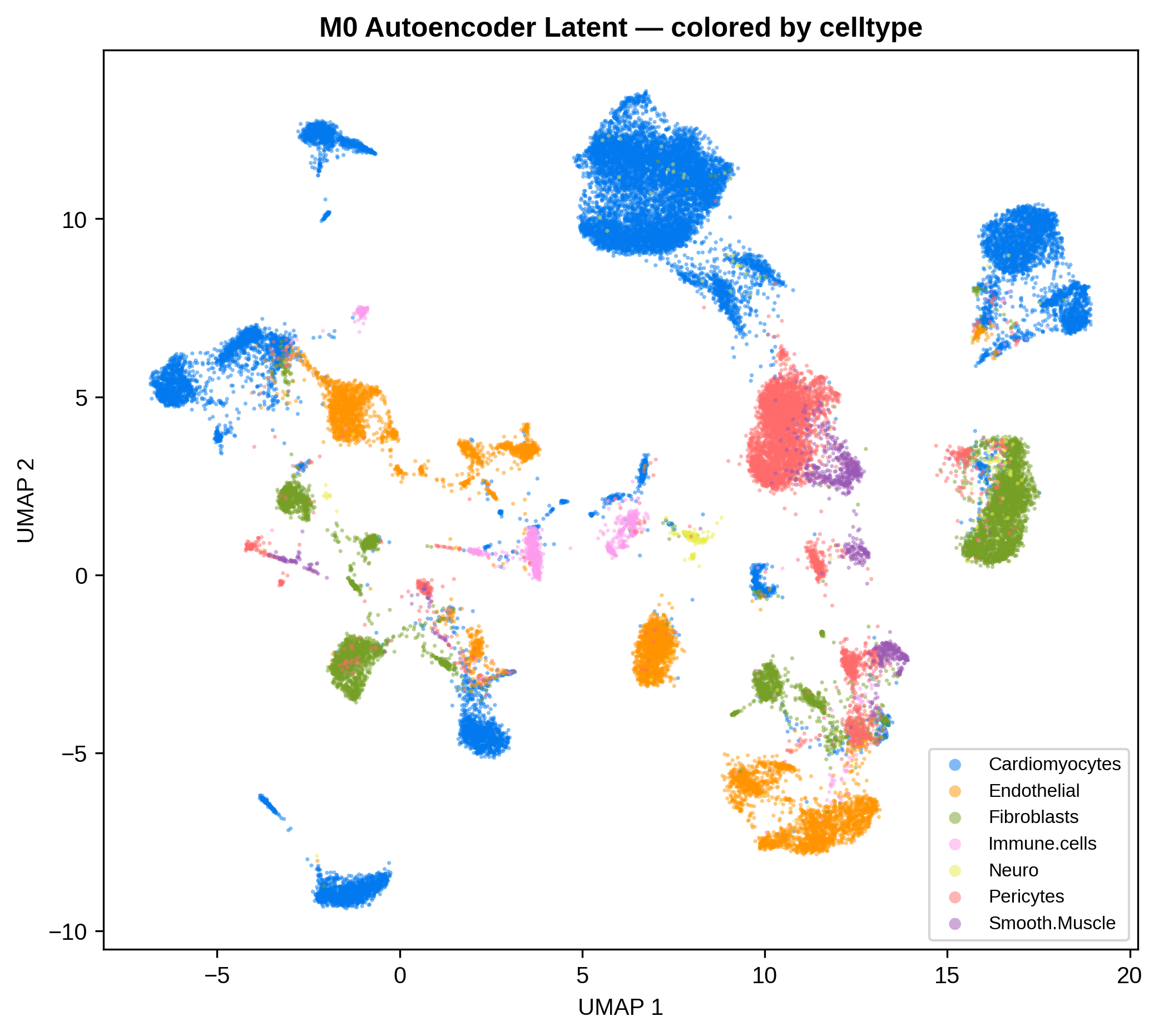}
    \end{minipage}\hfill
    \begin{minipage}{0.49\textwidth}
        \includegraphics[width=\textwidth]{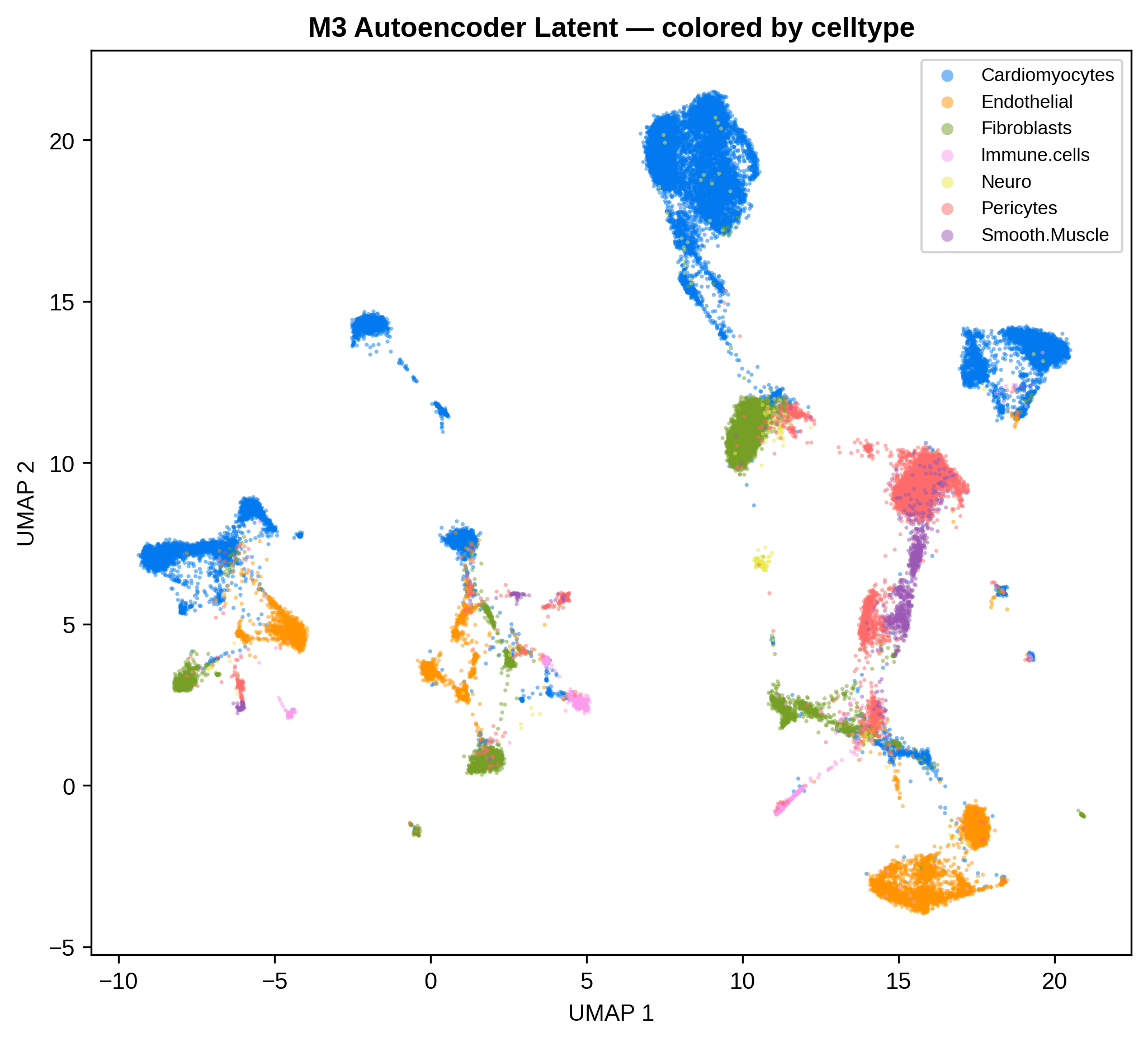}
    \end{minipage}
    \begin{minipage}{0.49\textwidth}
        \includegraphics[width=\textwidth]{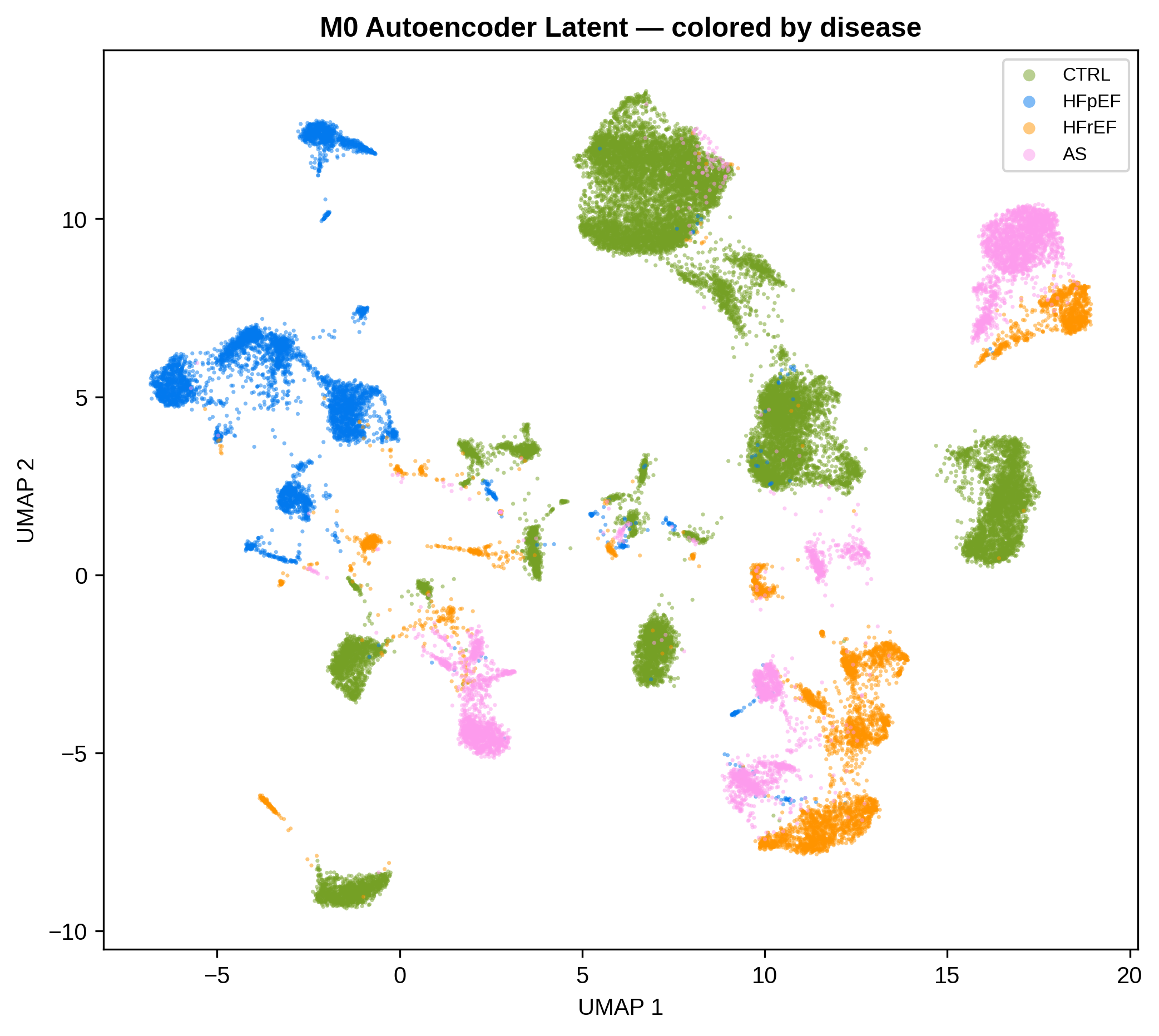}
    \end{minipage}\hfill
    \begin{minipage}{0.49\textwidth}
        \includegraphics[width=\textwidth]{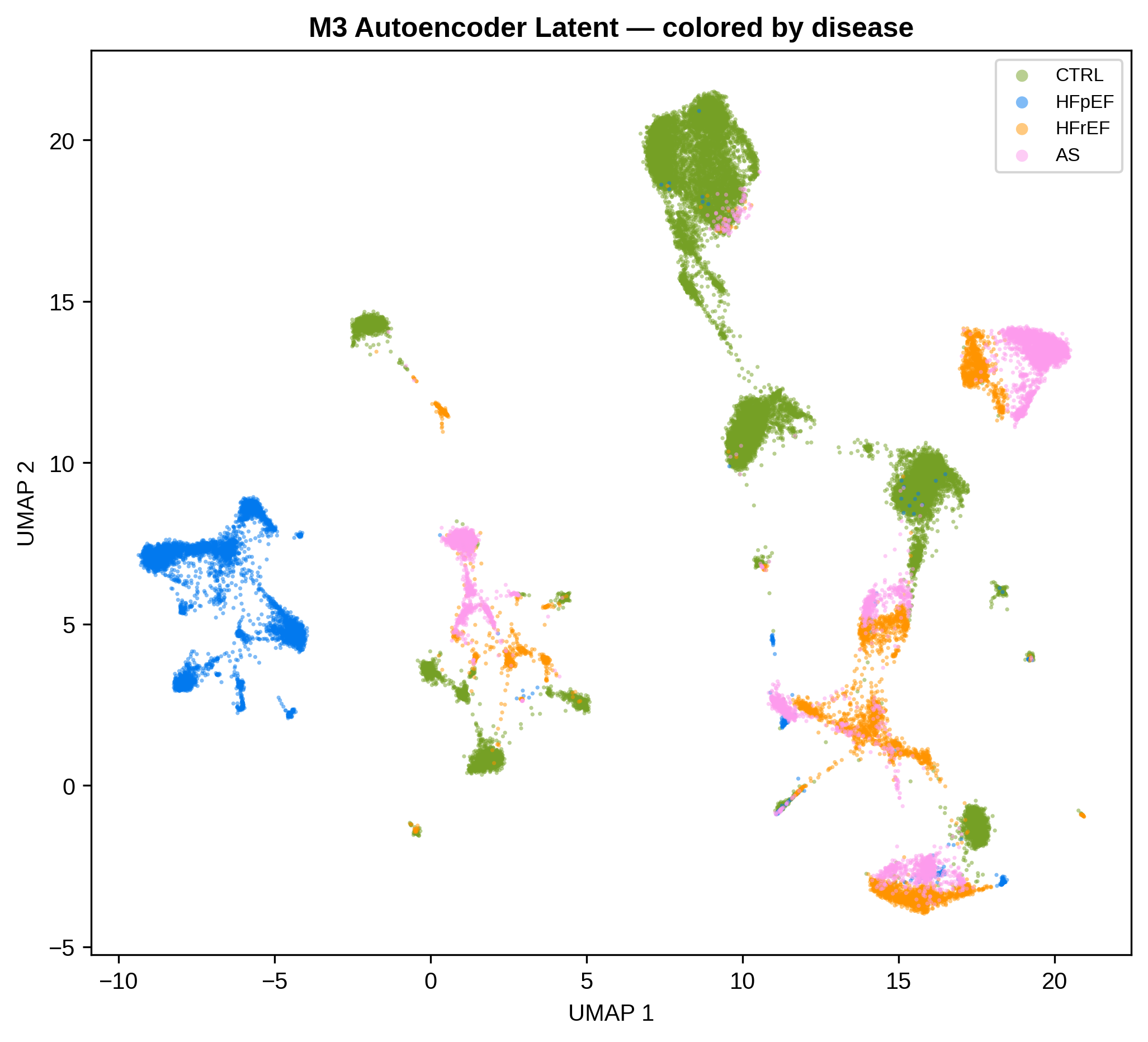}
    \end{minipage}
    \caption{UMAP projections of the 350-dimensional test-set representations. Left: dense DAE (M0). Right: BiMambaAE (M3). Top: nuclei colored by cell type. Bottom: nuclei colored by disease state. Both representations show cell-type structure despite the approximately $14.6\times$ difference in total autoencoder parameter count.}
    \label{fig:umap}
\end{figure}

UMAP projections \cite{mcinnes2018umap} of the test-set representations (Fig.~\ref{fig:umap}) suggest that both autoencoders separate major cell types and retain some within-cell-type structure associated with disease state. This qualitative visualization is consistent with the classification results, but does not establish preservation of the full latent geometry or quantify biological fidelity.

\subsection{DXG-based enrichment shows higher average agreement with the curated pathway reference}

\begin{figure}[t]
    \centering
    \begin{minipage}{0.49\textwidth}
        \includegraphics[width=\textwidth]{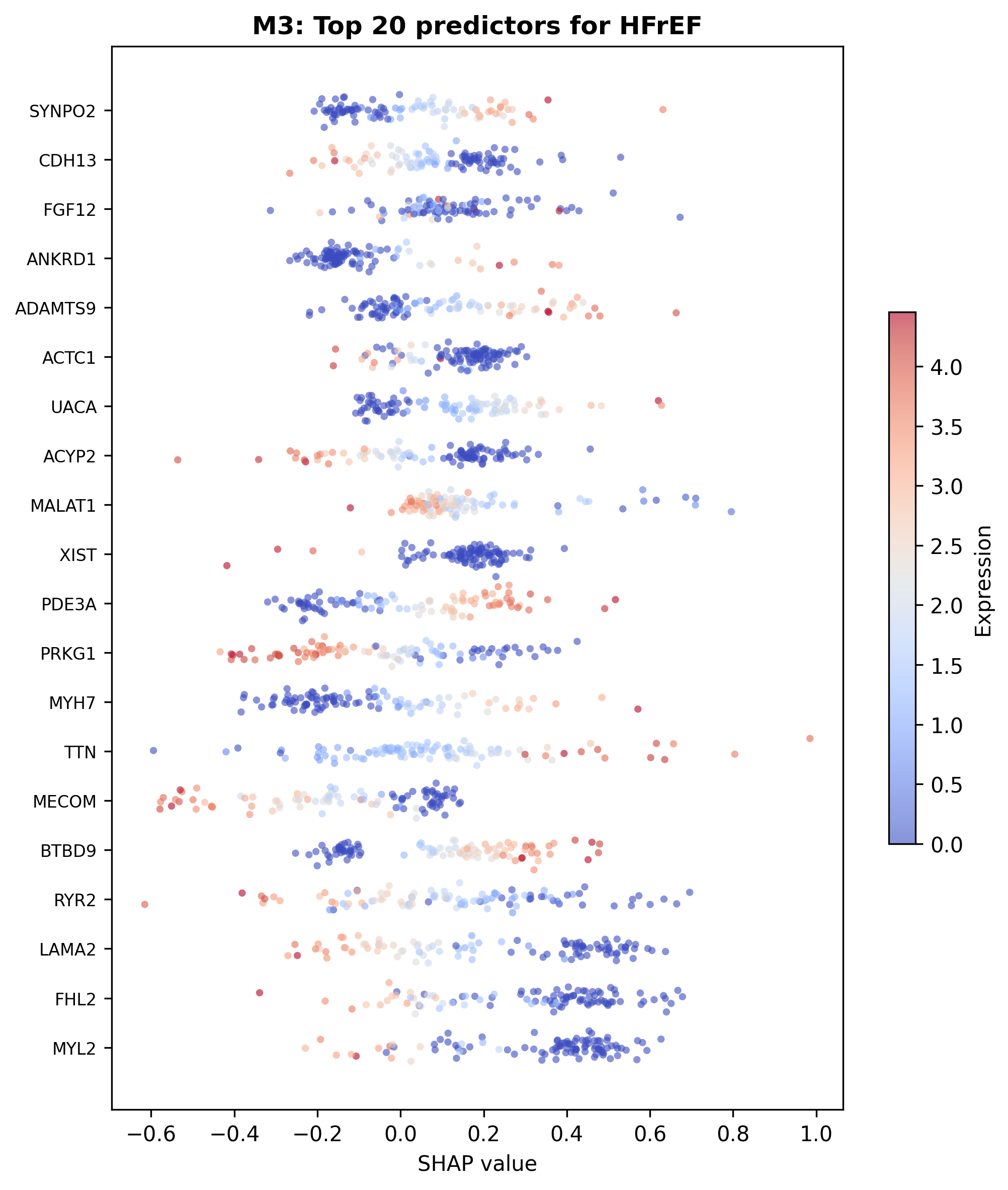}
    \end{minipage}\hfill
    \begin{minipage}{0.49\textwidth}
        \includegraphics[width=\textwidth]{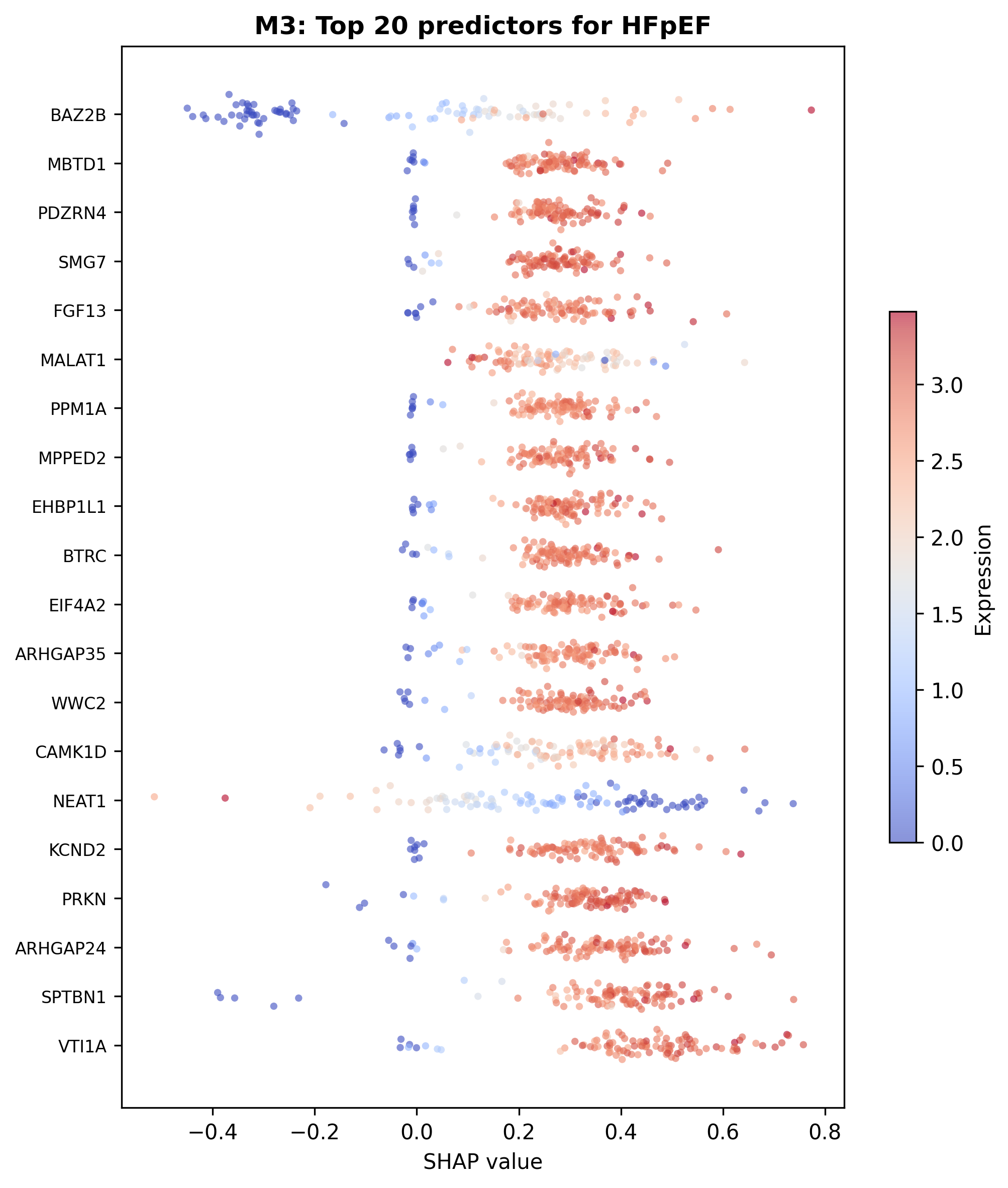}
    \end{minipage}
    \caption{SHAP beeswarm plots for the 20 highest-ranked gene attributions to the HFrEF (left) and HFpEF (right) outputs of M3 (BiMambaAE+MLP). Point color represents normalized expression. Highly attributed HFrEF genes include \emph{MYH7}, \emph{TTN}, \emph{RYR2}, \emph{ANKRD1}, \emph{ACTC1}, and \emph{MYL2}; attribution magnitude reflects contribution to model predictions rather than experimental validation of disease mechanisms.}
    \label{fig:beeswarm}
\end{figure}

\begin{figure}[t]
    \centering
    \includegraphics[width=0.55\textwidth]{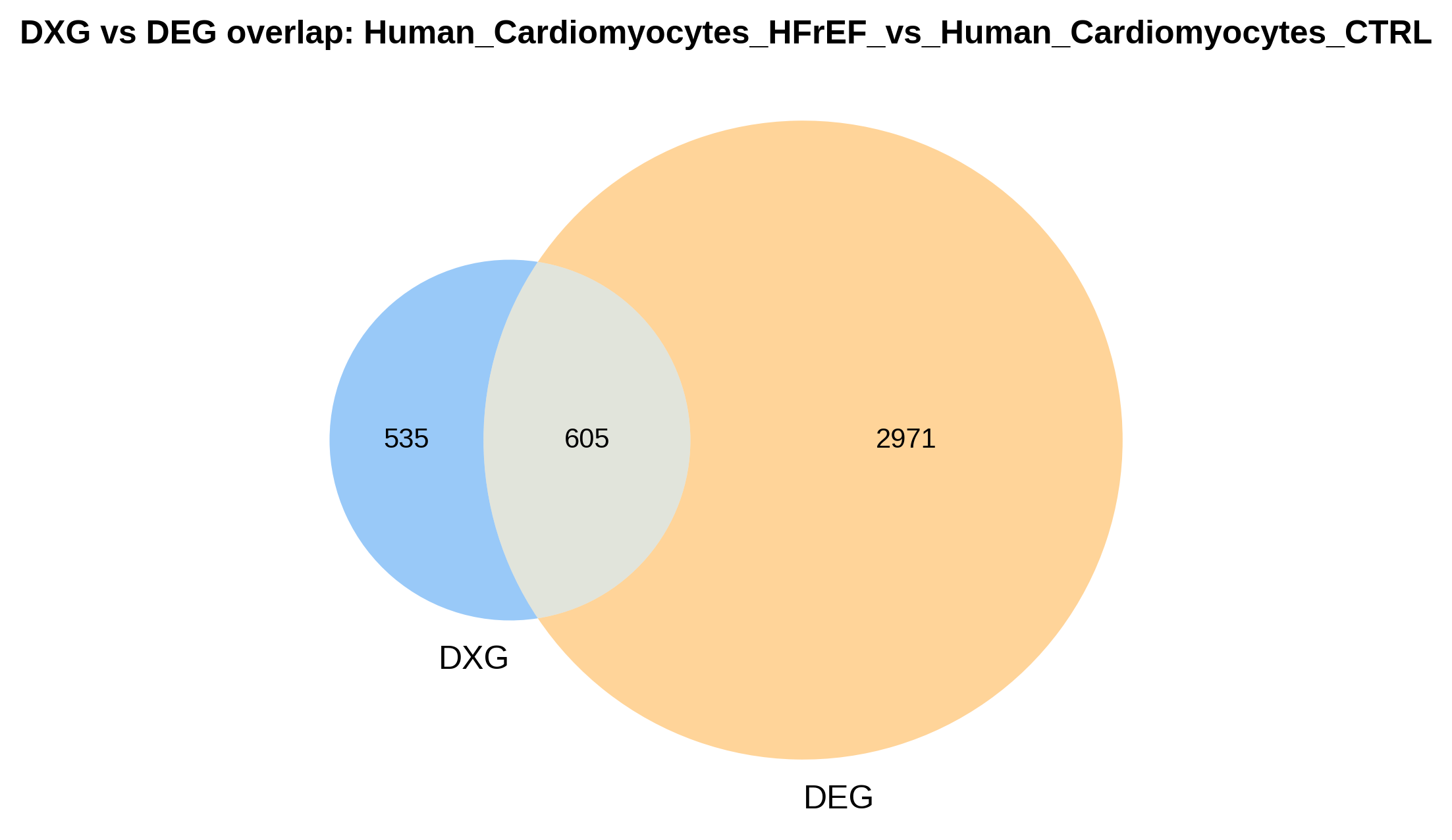}
    \caption{Overlap between M3 DXGs (SHAP-based) and DEGs (expression-based Wilcoxon test) for human cardiomyocytes in HFrEF vs.\ CTRL. The sets contain 1051 DXG-exclusive genes, 2807 DEG-exclusive genes, and 769 shared genes.}
    \label{fig:venn}
\end{figure}

SHAP analysis identified genes contributing to disease-state predictions. For HFrEF, highly attributed genes in M3 included cardiomyopathy- and remodeling-related loci such as \emph{MYH7}, \emph{TTN}, \emph{RYR2}, \emph{ANKRD1}, \emph{ACTC1}, \emph{MYL2}, \emph{FHL2}, \emph{LAMA2}, and \emph{PDE3A} (Fig.~\ref{fig:beeswarm}), consistent with the reference study \cite{jurado2025}. HFpEF attributions highlighted a different set, including \emph{MALAT1}, \emph{NEAT1}, \emph{KCND2}, \emph{CAMK1D}, and \emph{PRKN}. DXG analysis yielded 51{,}127 gene--contrast calls for M3 across 33 species$\times$cell-type$\times$condition contrasts, compared with 37{,}572 for M0. These totals count calls across contrasts rather than unique genes. For human cardiomyocytes in HFrEF vs.\ CTRL, 1820 DXGs and 3576 DEGs shared 769 genes (Fig.~\ref{fig:venn}). DXG-exclusive genes included \emph{DMD}, \emph{SORBS2}, \emph{MLIP}, \emph{PLCL1}, and \emph{ANK3}. These attributions describe model behavior and do not establish causal involvement in disease.

\begin{figure}[t]
    \centering
    \includegraphics[width=\textwidth]{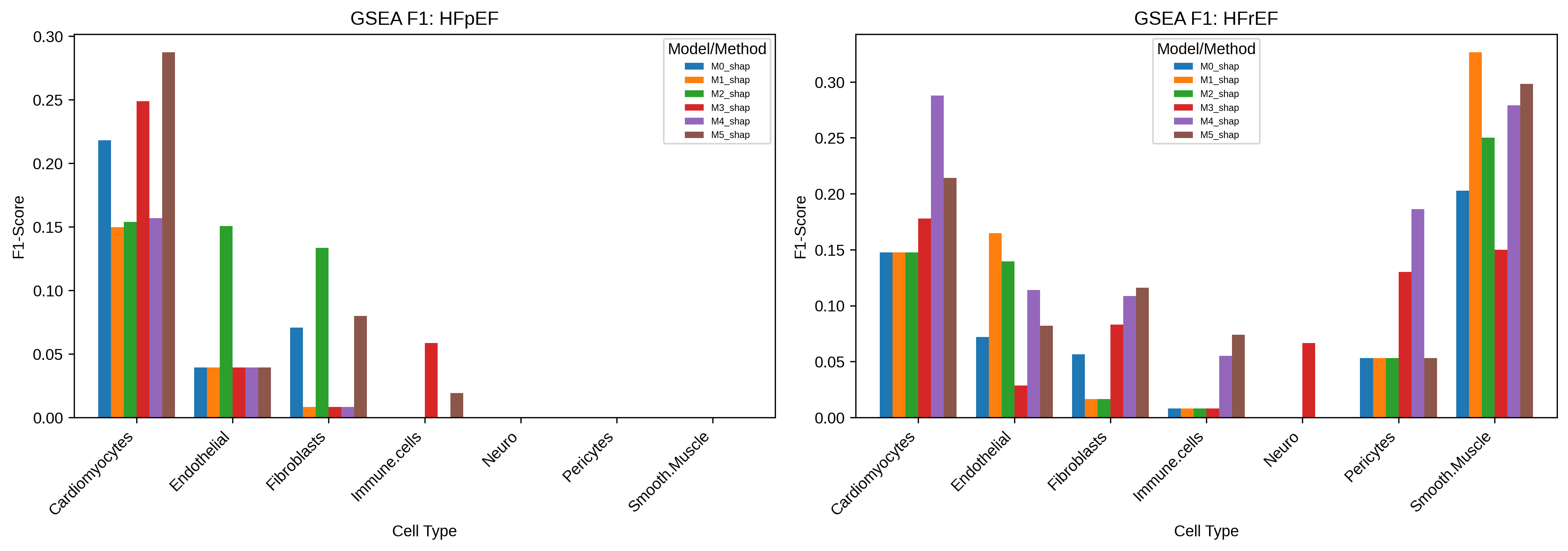}
    \caption{GSEA F1 scores against the curated, cell-type-specific GO keyword reference for HFpEF (left) and HFrEF (right). DXG-based rankings from each model are compared with DEG-based rankings. Scores describe agreement with the selected reference and vary across cell types and disease contrasts.}
    \label{fig:gsea}
\end{figure}

Against the curated GO keyword reference (Fig.~\ref{fig:gsea}), DXG-based GSEA achieved approximately twice the mean per-contrast F1 of DEG-based GSEA: 0.105 for M3-DXG over its 33 available contrasts vs.\ 0.053 for DEG over its 37 available contrasts. A sensitivity analysis restricted to the 31 contrasts available to both methods preserves the direction of the difference (0.112 vs.\ 0.051). Mean scores ranged from 0.095--0.108 for the three BiMambaAE configurations and from 0.076--0.086 for the DAE-based configurations. Examples favoring DXGs included mouse CM in HFpEF (F1 0.385 vs.\ 0.177), mouse IC in AS (0.500 vs.\ 0.000), mouse EC in AS (0.400 vs.\ 0.000), human FB in HFrEF (0.267 vs.\ 0.066), and human NC in HFrEF (0.200 vs.\ 0.000). The advantage was not uniform: DEGs achieved a higher score for human CM in HFrEF (0.310 vs.\ 0.091). The low absolute mean scores and variation across contrasts support interpreting DXGs as a complementary source of candidate pathways. Of the 42 possible contrasts, 39 were evaluable; the three unevaluable contrasts (human SMC in HFpEF, mouse NC in AS, mouse SMC in HFrEF) are flagged in Supplementary File S1.

\section{Discussion}

The results support the use of a compact bidirectional state-space autoencoder within this cardiac transcriptomic classification pipeline. BiMambaAE reduces the complete autoencoder parameter count by approximately $14.6\times$, while the BiMambaAE+MLP configuration improves the prediction metrics relative to the replicated dense DAE+MLP baseline. A possible explanation is that the compact latent-token representation provides a useful constraint on feature learning before input-dependent sequence mixing. However, the present comparisons do not distinguish this effect from changes in decoder capacity, optimization, batch size, or training duration. Dedicated component ablations under matched training conditions are needed to attribute the gains specifically to bidirectionality or state-space mixing. The reported species F1 of 1.000 and improvements in cell-type classification characterize this held-out test set and do not establish generalization to independent cohorts.

Classifier choice remained consequential. MLP and XGBoost produced similar scores on frozen BiMambaAE representations, whereas the graph classifiers and unweighted BCE-trained SSM heads performed less well. The improved scores obtained with class weighting or attention pooling indicate that SSM-head performance is sensitive to training and aggregation choices in this setting. The MLP offers a practical combination of predictive performance, low measured classifier latency (0.15~ms/batch), and differentiability with respect to the original expression inputs. This final property enables the gradient-based attribution workflow used for DXG analysis.

The pathway analysis extends the DXG evaluation of Ruz Jurado et al.\ \cite{jurado2025} to a state-space autoencoder. The approximately twofold increase in average reference-based GSEA F1 relative to DEGs suggests that prediction-oriented attributions can prioritize biological information not captured by expression differences alone. Nevertheless, attribution identifies features used by a fitted model; it does not independently validate their mechanistic relevance. DXG-exclusive genes, including \emph{DMD} in HFrEF cardiomyocytes, provide candidates for follow-up investigation. Because DEGs perform better in some contrasts, both analyses should be reported, and any benefit from combining their gene sets should be tested explicitly.

\paragraph{Relationship to single-cell foundation models.} Geneformer, UCE, and scFoundation target broadly transferable representations learned from large pretraining corpora \cite{theodoris2023transfer,rosen2026uce,hao2024scfoundation}. scHFBiMamba instead evaluates a task-specific autoencoder within a defined cardiac classification and interpretation workflow. Its 13.2M-parameter autoencoder and differentiable classifier provide a compact implementation for this setting. No foundation-model baseline was evaluated, so the present results do not establish advantages in accuracy, transferability, computational cost, or attribution quality over pretrained alternatives. Future work could assess frozen and fine-tuned foundation-model embeddings under the same sample-level split and downstream evaluation protocol.

\paragraph{Limitations.}

Experiments were restricted to one two-species dataset. Independent cohort validation, repeated training runs, and evaluation of sensitivity to data partitioning are needed to characterize robustness. Although test samples were held out at the biological-sample level, nucleus-level metrics and significance tests do not account for dependence among nuclei from the same donor or animal. Sample-aware uncertainty estimates would therefore strengthen inference. The reported DXG and DEG tests also require careful interpretation because their observational units are cells rather than independent biological replicates. Further limitations concern the keyword-based enrichment reference, implementation-specific efficiency measurements, and incomplete isolation of architectural effects.

External validation could include an independent dilated-cardiomyopathy atlas such as that of Koenig et al.\ \cite{koenig2022single}, provided that source-sample overlap is excluded. Most configurations were trained with one seed and one training/validation split, and the classifier comparison reports single runs; its rows derive from different runs and latent spaces, as disclosed in the Results, which limits direct comparability across tables. The GSEA evaluation depends on curated reference keywords, produces modest absolute F1 values, and favors DXGs inconsistently across contrasts. The current BiMambaAE implementation depends on CUDA kernels from \texttt{mamba\_ssm}, limiting CPU-only deployment. Token count, layer depth, and state size were selected heuristically and warrant systematic sensitivity analysis. The GCN/GAT FLOPs and timings were measured with a hidden-128 benchmark configuration rather than the trained hidden-64 models (Table~\ref{tab:efficiency}), and benchmark hardware and software versions are not archived with the results bundle. Comparisons with pretrained single-cell foundation models remain future work.

\section{Conclusion}

scHFBiMamba provides a compact autoencoder-based framework for classifying cardiac nuclei and analyzing gene-level model attributions. In the reported experiments, BiMambaAE with an MLP head achieved higher classification scores than the dense DAE+MLP comparator while reducing the total autoencoder parameter count by approximately $14.6\times$. DXG-derived rankings showed higher average agreement with a curated HF-related pathway reference than DEG-derived rankings, with substantial variation across contrasts. These findings motivate further evaluation using matched training controls, sample-aware statistical analysis, and independent cohorts.

\section*{Data and code availability}
The CVD dataset is publicly available at \url{https://huggingface.co/datasets/mruzjurado/CVD_dataset} \cite{cvd_dataset}. The scHFBiMamba code is available at \url{https://github.com/UgoLomoio/scHFBiMamba}.

\section*{Acknowledgments}
The authors thank Ruz Jurado et al.\ for making the CVD dataset and analysis code publicly available. This work was supported by the OFIDIAPlus project, INTERREG GREECE--ITALY 2021--2027 PROGRAMME. All scientific content, results, and interpretations were verified, modified and approved by the authors. 

\section*{Declaration of interests}
The authors declare that they have no known competing financial interests or personal relationships that could have appeared to influence the work reported in this paper.

\subsection*{Ethics statement}
This study did not require ethical approval because it used a publicly available, de-identified dataset and did not involve direct interaction with human subjects.

\clearpage
\section*{Supplementary Material}

\begin{figure}[h]
    \centering
    \includegraphics[width=\textwidth]{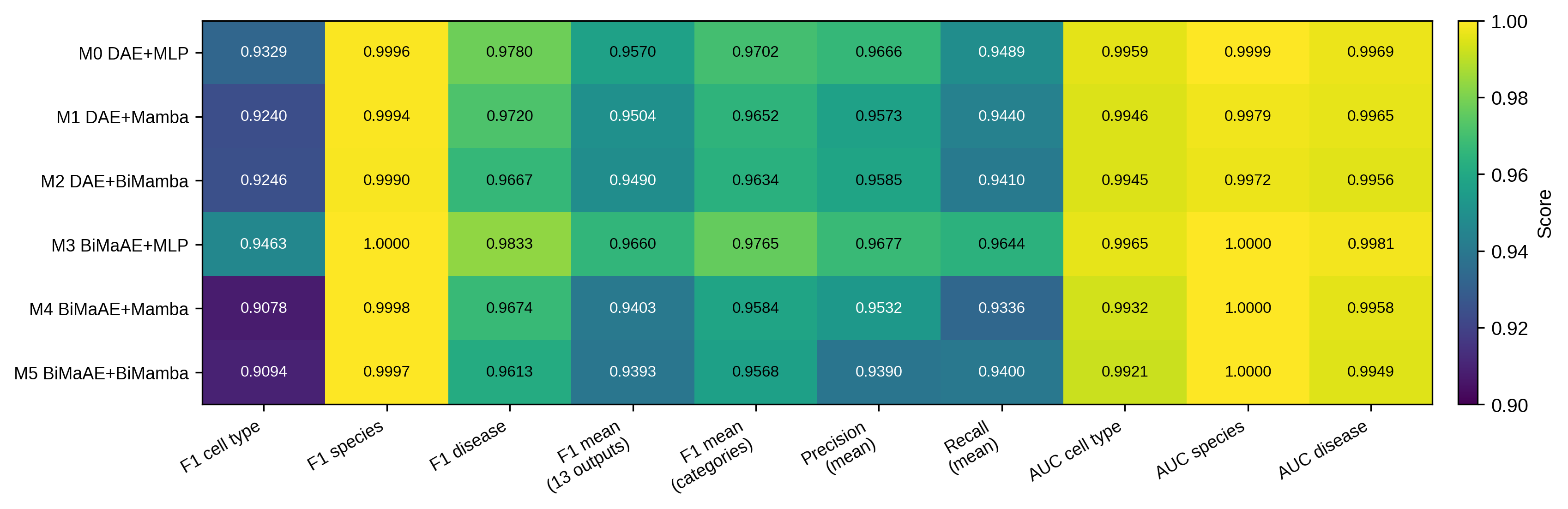}
    \caption{Supplementary Figure S1: Ablation heatmap across the six configurations (rows) and ten performance metrics (columns): macro-F1 per category, both aggregate F1 statistics (``F1 mean (13 outputs)'' and ``F1 mean (categories)''), mean precision and recall, and macro ROC-AUC per category, all on the held-out test set.}
    \label{fig:ablation_heatmap}
\end{figure}

\begin{figure}[h]
    \centering
    \begin{minipage}{\textwidth}
        \includegraphics[width=\textwidth]{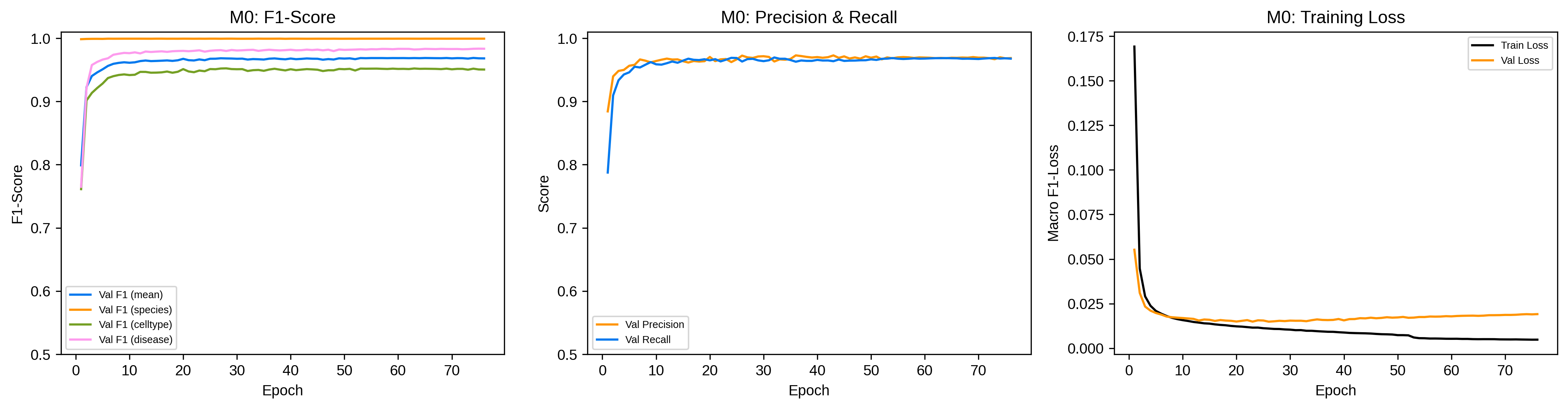}
    \end{minipage}\hfill
    \begin{minipage}{\textwidth}
        \includegraphics[width=\textwidth]{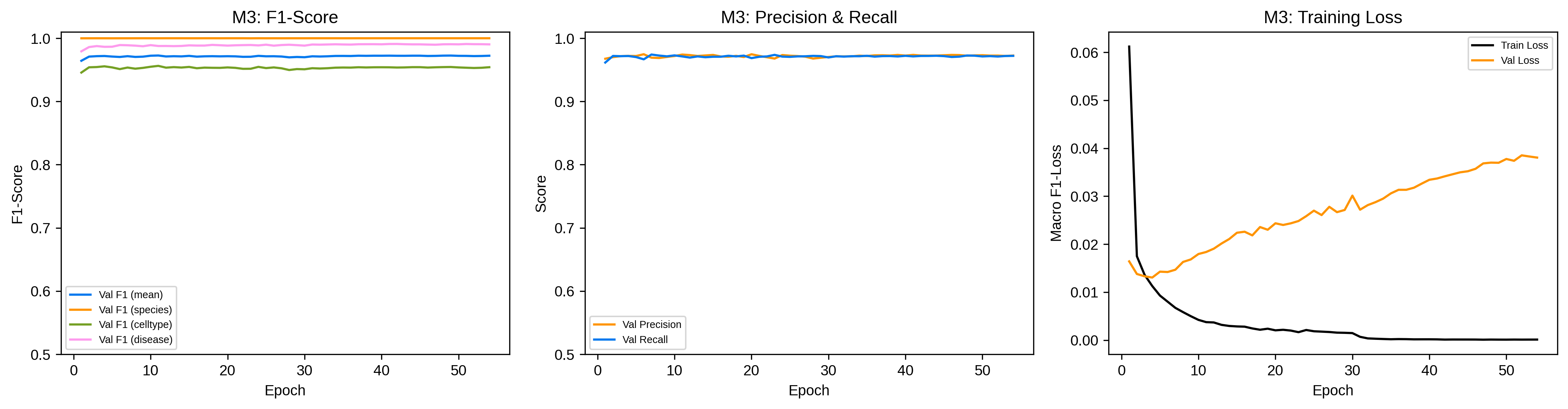}
    \end{minipage}
    \caption{Supplementary Figure S2: Training and validation curves for the replicated baseline M0 (left) and the proposed M3 (BiMambaAE+MLP, right).}
    \label{fig:training_curves}
\end{figure}

%Bibliography
\bibliographystyle{unsrt}
\bibliography{references}

\end{document}